\pdfoutput=1
\providecommand{\projecturl}{\url{https://hhjj233.github.io/CornerPercentile/}}
\PassOptionsToPackage{hyphens}{url}
\documentclass[a4paper,fleqn]{cas-sc}
\usepackage[authoryear]{natbib}

\renewcommand{\ttdefault}{txtt}
\usepackage{graphicx}
\usepackage[labelsep=period]{caption}
\usepackage{subcaption}
\usepackage{booktabs}
\usepackage{tabularx}
\usepackage{amsfonts}
\usepackage{amsmath,amssymb,amsthm}
\usepackage{microtype}
\usepackage{stfloats}

\usepackage{algorithm}
\usepackage{algpseudocode}

\newcommand{\papertitle}{How corner is a corner case? Percentile control for highway scenario generation}
\AddToHook{env/Abstract/begin}{\hfuzz=\dimexpr.25\textwidth+1pt\relax}
\ExplSyntaxOn
\keys_define:nn { cas / fig }
  { unknown .code:n = { \tl_if_blank:nT {#1} { \tl_set:Nx \l_fig_pos_tl { \l_keys_key_str } } } }
\keys_define:nn { cas / tbl }
  { unknown .code:n = { \tl_if_blank:nT {#1} { \tl_set:Nx \l_tbl_pos_tl { \l_keys_key_str } } } }
\ExplSyntaxOff

\newcommand{\E}{\mathbb{E}}
\newcommand{\Prob}{\mathbb{P}}

\newcommand{\Tcap}{\tau}

\newcommand{\Fine}{\mathrm{Fine}_{0.05}}
\newcommand{\G}{\widehat F}
\newcommand{\ind}{\mathbf{1}}
\begin{document}
\let\WriteBookmarks\relax
\def\floatpagepagefraction{1}
\def\textpagefraction{.001}
\shorttitle{Percentile control for highway scenario generation}
\shortauthors{J. Liu et~al.}

\title[mode=title]{\papertitle}

\author[1]{Jiaxi Liu}
\author[1]{Hang Zhou}
\author[1]{Hangyu Li}
\author[1]{Yifan Wang}
\author[1]{Keke Long}
\author[1]{Chengyuan Ma}
\author[1]{Bin Ran}
\author[1]{Xiaopeng Li}
\cormark[1]
\ead{xli2485@wisc.edu}
\affiliation[1]{organization={Department of Civil and Environmental Engineering, University of Wisconsin--Madison},
    city={Madison},
    postcode={53706},
    state={WI},
    country={USA}}
\cortext[cor1]{Corresponding author}

\begin{abstract}
Generating corner-case scenarios with appropriate adversity in a simulation environment is critical for testing an autonomous vehicle (AV) software stack’s safety performance before deployment. Existing autonomous-driving scenario generators can enforce specific behavior, adversity, or feasibility conditions, but they provide limited control over how extreme a generated scenario is relative to plausible futures in the same traffic context. This study represents the adversity of a generated scenario as its percentile in the conditional distribution of future risk given the observed history. This view supports calibrated answers to two questions: how “corner” a generated corner-case scenario is and how its “cornerness” can be fine-tuned.

To this end, we formulate history-conditioned risk-percentile requests and learn a reference risk distribution that maps each requested percentile to a physical risk target. We then use a percentile-conditioned joint diffusion model with sampling-time risk guidance to generate multi-agent futures, together with a reference-based criterion for evaluating percentile realization.

Experiments use the minimum post-encroachment time (PET) between the ego and its surrounding vehicles as the risk surrogate on highD. On the primary evaluation set, our method realizes 1,422 of 1,440 requests within a 0.05 percentile tolerance (98.75\%), with mean percentile error 0.00673 and PET-target error 0.00991 seconds. The resulting interface connects context-relative risk specification, physical realization, and evaluation through a common risk scale. Project website and videos of generated scenarios are available at \projecturl.
\end{abstract}
\begin{keywords}
Automated vehicles \sep Traffic scenario generation \sep Natural driving data \sep Conditional risk percentiles \sep Request semantics
\end{keywords}

\maketitle

\suppressfloats[t]

\begin{figure*}[t]
\centering
\begin{subfigure}[t]{.485\textwidth}
\centering
\includegraphics[width=\linewidth]{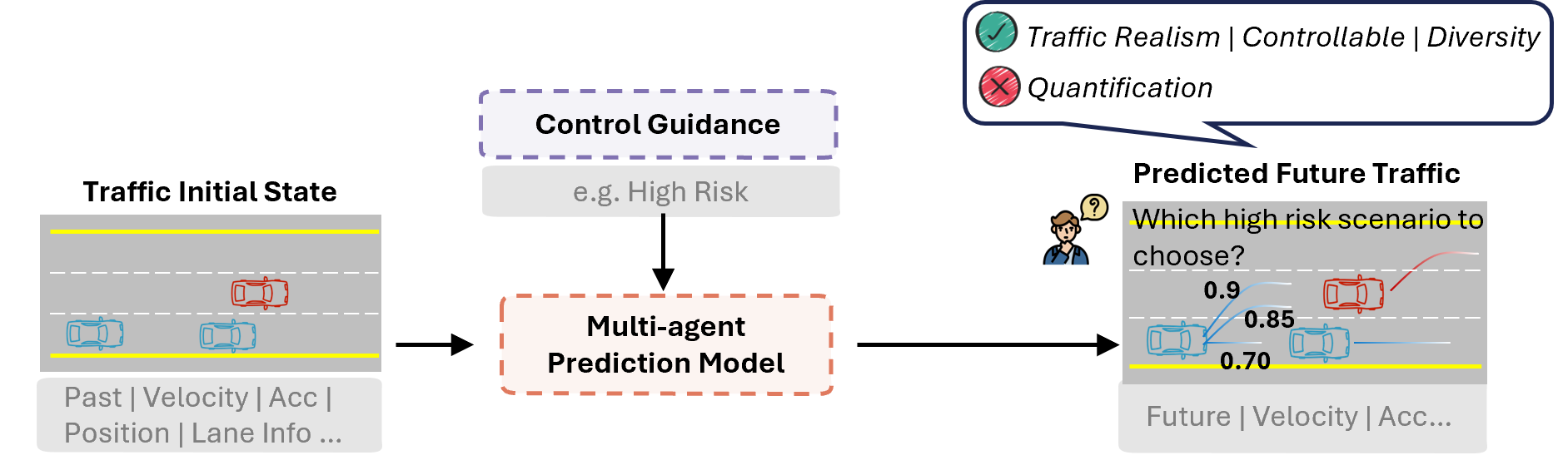}
\caption{Existing interfaces use tokens, language prompts, or constraints and leave adversity percentiles unquantified.}
\end{subfigure}\hfill
\begin{subfigure}[t]{.485\textwidth}
\centering
\includegraphics[width=\linewidth]{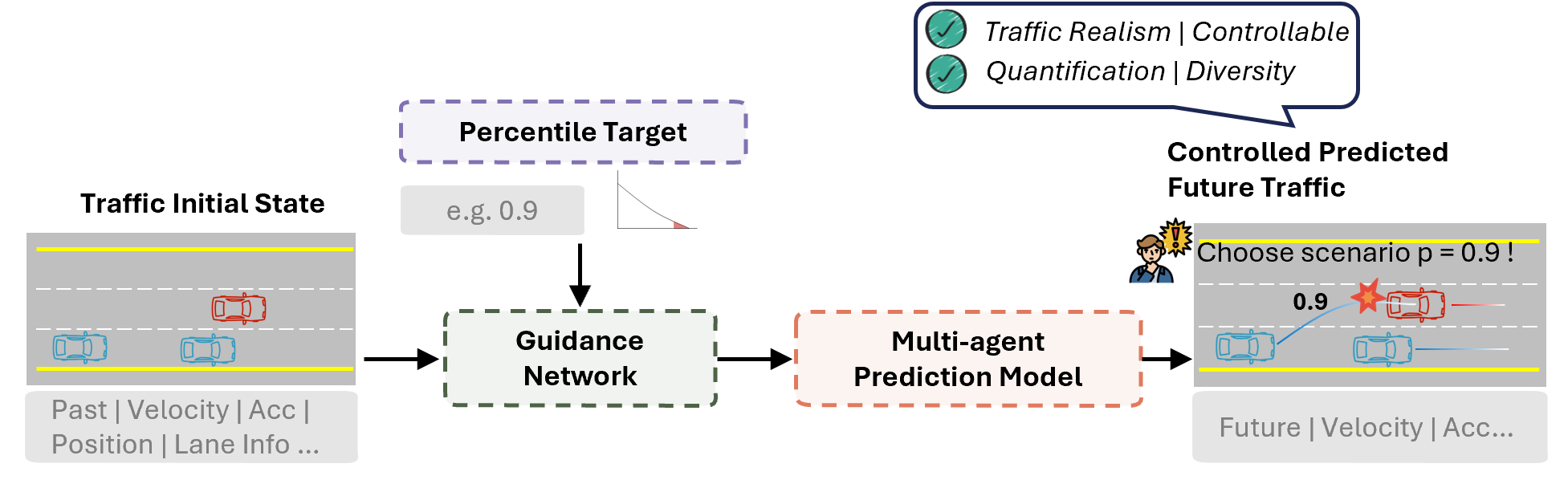}
\caption{Our view: corner cases are indexed by a target adversity percentile $p$.}
\end{subfigure}
\caption{Why adversity percentiles matter. Existing scenario-generation methods offer control variables that are usually uncalibrated to context-conditional adversity.}
\label{fig:three_vertical}
\end{figure*}

\section{Introduction}
Before deployment, the software stack of an autonomous vehicle (AV) is always tested on corner-case scenarios whose adversity suits the test at hand. Scenario-based testing organizes traffic situations into repeatable cases for this purpose \citep{menzel2018scenarios,riedmaier2020survey}, and a simulation environment can generate such cases on demand. A useful scenario library spans ordinary, moderately critical and highly critical futures, so the generator must also state how extreme each generated scenario is.

Existing autonomous-driving scenario generators can already enforce behavior, adversity or feasibility conditions. Rare-event sampling and stress testing seek informative or adverse outcomes \citep{okelly2018rare,koren2018ast}, while adversarial trajectory generation and learned traffic priors expand the interactions available for testing \citep{wang2021advsim,rempe2022strive,zhang2023cat}. Diffusion models support motion constraints and scene-level requests \citep{jiang2023motiondiffuser,zhong2023guided,ctgpp2023}, and language-conditioned methods express desired scene semantics \citep{tan2023lctgen,zhou2026enhancing}. Risk indices derived from post-encroachment time (PET) are used in risk-aware trajectory models \citep{wang2025rade}, and risk conditions guide ego planning \citep{qian2026riskdiffuser}. These controls specify what a scenario should contain, but they offer limited control over how extreme the result is relative to the plausible futures of the same traffic context, as shown in Fig.~\ref{fig:three_vertical}.

To control adversity, we first note that adversity is context-relative. The same one-second PET can occupy different positions among natural traffic futures. For a history whose futures usually maintain wide temporal separation, it lies toward the critical tail. For another history whose futures typically pass more closely, it can lie near the center of the distribution. We therefore represent the adversity of a generated scenario as its percentile in the conditional distribution of future risk given the observed history. This one calibrated scale answers two questions: how ``corner'' a generated corner-case scenario is, and how its ``cornerness'' can be fine-tuned by requesting a different percentile. A request of this kind must determine both the physical outcome to pursue and the criterion for recognizing its realization.

A risk surrogate specifies the interaction property to be ranked, and a history-conditioned natural distribution supplies the reference for that ranking. Criticality measures capture geometric, temporal and behavioral aspects of encounters \citep{westhofen2023criticality,lin2023commonroad}, while natural-data risk models use interaction-dependent distributions to interpret traffic conditions \citep{jiao2026learning}. We make a position in such a distribution the generation request itself. Let $\widehat F_H$ denote the estimated natural conditional cumulative distribution function (CDF) of a scalar surrogate for history $H$, oriented so that smaller values mean greater criticality. We call $\widehat F_H$ the reference risk distribution. The requested risk percentile $p$ determines the physical risk target $q_H(p)=\widehat F_H^{-1}(1-p)$ through the generalized inverse. Thus $p=0.9$ selects the 0.1 quantile of the surrogate distribution, and holding $p$ fixed across histories allows the physical target to adapt to each context. PET is the surrogate used in this study, and the request is defined by the conditional position of the selected scalar measure.

To generate such scenarios, we learn the reference risk distribution from natural driving observations and condition a joint diffusion model on the request. Each recorded history usually supplies one realized future, so we pool supervision across history--future pairs and learn the conditional distribution with the continuous ranked probability score (CRPS). A temporal--relational reference reads the historical interactions between the ego and its surrounding vehicles across the PET range. Once fitted and calibrated, reference models supply percentile labels for training and translate each request into its physical risk target at generation time. The percentile-conditioned joint diffusion model receives $p$, its physical risk target and a CDF descriptor, and sampling-time risk guidance moves the physical outcome toward that target. Masked actor attention accommodates varying vehicle counts, and road and background-separation constraints act within the same sampling path, so the generated multi-agent futures remain physically consistent.

The task and request interface form the central contribution. The implementation and decomposed evidence establish how the interface can be used:
\begin{enumerate}
\item We represent the adversity of a generated scenario as a history-conditioned risk percentile and define a request as a specified position in the natural future-risk distribution. A reference risk distribution maps the request to a physical risk target, and a reference-based criterion evaluates its realization, giving the request one consistent meaning from specification to evaluation.
\item We provide an executable implementation: a calibrated reference risk distribution learned from natural driving data and a percentile-conditioned joint diffusion model whose sampling-time risk guidance links the requested percentile to a physical multi-vehicle future.
\item We provide decomposed evidence for reference prediction, request realization and the roles of conditioning and guidance, together with same-history analyses of external prior samples.
\end{enumerate}
\section{Related work}
\subsection{Traffic generation and control interfaces}
Traffic-generation research has established several ways to specify desired outcomes \citep{ding2023survey}. Rare-event estimation targets informative cases, adaptive stress testing searches for likely failures, and probabilistic scene descriptions encode desired configurations \citep{sinha2020nbs,lee2020adaptive,fremont2019scenic}. For AV safety evaluation, importance sampling and testing scenario libraries concentrate tests on critical cases while keeping performance estimates unbiased \citep{zhao2017lanechange,feng2020tslg1,yang2025adaptive}. Learned background agents apply sparse adversarial adjustments to naturalistic traffic for the same purpose \citep{feng2021nade,feng2023d2rl}. Adversarial trajectory perturbations and learned traffic priors incorporate multi-vehicle interactions into scenario search \citep{wang2021advsim,rempe2022strive,zhang2026harm}. Joint diffusion models support motion and scene-level constraints \citep{jiang2023motiondiffuser,zhong2023guided,ctgpp2023}, and guidance weights or interpolated expert policies let users tune how adversarial the generated agents behave \citep{chang2024safesim,nie2026sage}. Language-conditioned methods express semantic and behavioral goals \citep{tan2023lctgen}. These contributions make desired scene properties available as generation inputs. Our focus is a distributional specification for one such input: the requested position of a risk outcome among natural futures conditioned on the observed history. The interface links that position to a physical target and evaluates the generated outcome against the same conditional reference.

\subsection{Risk surrogates and context-relative severity}
Traffic criticality measures summarize aspects of an encounter such as proximity, timing and behavioral response \citep{westhofen2023criticality,lin2023commonroad}. PET measures temporal separation through occupation of shared space \citep{fhwa2008surrogate}. RADE conditions joint multi-agent diffusion on a risk level computed from scene PET \citep{wang2025rade}. A conditional diffusion model guided by a large language model generates car-following and cut-in scenarios on highD at risk levels given by intervals of modified time to collision or PET \citep{zhou2026enhancing}, and RiskDiffuser conditions ego planning on a continuous risk level \citep{qian2026riskdiffuser}. These methods establish the value of supplying risk-related conditions to generation or planning, and each expresses the condition on a fixed physical scale. The formulation closest to ours is that of \citet{zhou2025quantile}, who generate scenarios at a specified quantile of a risk index by searching a parametric car-following scenario space in simulation with importance sampling and particle swarm optimization. Their quantile refers to the distribution of the risk index over that scenario space, whereas our percentile refers to the natural futures of one observed history. Natural spacing distributions provide context-dependent references for interpreting interaction outcomes \citep{jiao2026learning}. We bring this reference into the request's definition: $p$ selects a position in the estimated natural conditional distribution, and its generalized inverse supplies the physical risk target. The minimum PET between the ego and its surrounding vehicles (SVs) provides our experimental surrogate. The formulation assigns a relative position to the selected measure, with its criticality direction specified as part of the task.

\subsection{Conditional distributions and calibration}
Quantile regression and proper scoring rules support learning conditional outcome distributions \citep{koenker1978rq,gneiting2007strictly}. Distribution recalibration adjusts their predictive probabilities \citep{kuleshov2018accurate,song2019distribution,dheur2023large}, and conditional diagnostics examine reliability across contexts that pooled summaries can obscure \citep{zhao2021diagnostics,dey2025instancewise}. We use these statistical tools to estimate the natural reference that gives a percentile request its meaning. The implementation combines temporal history encoders with a set-attention readout \citep{lee2019set}, allowing each PET-distribution component to query historical ego--SV relations. Its coherent CDF supports percentile supervision, generalized-inverse target queries and compatible-interval outcome scoring. Predictive quality of this estimated reference and realization of its requests are evaluated as distinct parts of the interface.

\section{History-conditioned risk-percentile requests}
\label{sec:traffic-formulation}
\label{sec:problem}
\subsection{From a risk value to a contextual request}
A risk percentile specifies how critical, or how adverse, a generated future should be relative to natural traffic following the observed history. The same percentile can therefore correspond to different physical outcomes in different scenes. To make this request precise, let $H$ denote an observed traffic history together with static scene information, and let $X$ denote a joint future of the ego and its SVs. Choose a scalar surrogate $R(X,H)$ and specify its criticality direction. We orient it as
\begin{equation}
Y(X,H)=d\,R(X,H),\qquad d\in\{-1,1\},
\label{eq:oriented-surrogate}
\end{equation}
where $d$ is chosen so that smaller $Y$ means greater criticality. Thus $d=1$ for a smaller-is-worse time gap, whereas $d=-1$ reverses a larger-is-worse surrogate. The application determines the surrogate, and the percentile expresses its conditional ordering.

This ordering requires a reference for what is natural after the current history. Let $F^\star_H$ be the conditional CDF of $Y$ in the chosen natural reference population,
\begin{equation}
F^\star_H(y)=\Prob_{\rm ref}\{Y(X,H)\le y\mid H\}.
\label{eq:population}
\end{equation}
We estimate this distribution from natural history--future pairs and denote its calibrated estimate by $\G_H$. The reference gives a raw surrogate value its contextual meaning: the same $y$ can occupy different conditional ranks for two different $H$.

A request $p\in(0,1)$ selects a risk percentile, with larger values indicating greater criticality. For example, $p=0.9$ addresses the lower tenth of the conditional surrogate distribution. Because $Y$ is oriented smaller-is-worse, we translate the request into the physical risk target
\begin{equation}
q_H(p)=\inf\{y:\G_H(y)\ge 1-p\}.
\label{eq:percentile-target}
\end{equation}
Holding $p$ fixed preserves the requested relative criticality while allowing the numerical target to change with $H$. The generator maps $(H,p,z)$ to $X^{\rm gen}=G_\theta(H,p,z)$, where $z$ controls sampling variability, and seeks to realize $Y(X^{\rm gen},H)\approx q_H(p)$. The request thus expresses a position in the selected scene-level surrogate distribution.

\subsection{Evaluating a percentile request consistently}
The inverse-CDF target also determines how realization should be evaluated. A surrogate distribution can contain a point mass (an atom), for example at a zero time gap or at a time gap clipped at a safe-end cap. Several percentile requests then share one physical target. We therefore associate an outcome $y$ with its compatible percentile interval,
\begin{equation}
I_H(y)=[1-\G_H(y),\,1-\G_H(y^-)].
\label{eq:rank}
\end{equation}
At a continuous point this interval reduces to the usual percentile $1-\G_H(y)$. At a point mass, every $p$ in the interval is compatible with the same physical outcome.

We measure how far the requested percentile lies from the interval realized by the generated future. For $y_{\rm gen}=Y(X^{\rm gen},H)$, we define the interval error $e_{\rm int}$ and two summaries,
\begin{align}
e_{\rm int}(H,p,z)&=\operatorname{dist}(p,I_H(y_{\rm gen})),\\
\mathrm{P\text{-}MAE}&=\E[e_{\rm int}],\\
\Fine&=\E[\ind\{e_{\rm int}\le 0.05\}].
\label{eq:risk-mae-goal}
\end{align}
P-MAE is the mean percentile error. The fine-control rate $\Fine$ is the fraction of requests realized within 0.05 in percentile. For $p=0.9$ and a continuous outcome, this means a generated risk percentile between 0.85 and 0.95.
The generalized inverse satisfies
\begin{equation}
\G_H(q_H(p)^-)\le 1-p\le\G_H(q_H(p)),
\label{eq:quantile-compatibility}
\end{equation}
so exact realization of the physical risk target has zero interval error, including at atoms. Physical-target MAE, $\E[|y_{\rm gen}-q_H(p)|]$, complements this measure by preserving the surrogate's physical resolution. We report both and separate atom-target from continuous-target requests.

\subsection{PET instantiation for a multi-vehicle future}
Our experiments instantiate $Y$ as minimum ego--SV occupancy PET, so $d=1$: a smaller temporal gap between shared-space occupancies indicates a more critical encounter. Let $\mathcal A(H)$ be the SVs defined from the observed history, and $B_i(t)$ a vehicle's footprint during the future window $W=[0,6.96]$ seconds. With measured dimensions and piecewise-linear positions between native frames,
\begin{align}
Y_{ej}(X,H)&=\inf_{\substack{s,t\in W\\B_e(s)\cap B_j(t)\ne\varnothing}}|s-t|,\\
Y(X,H)&=\min\{\Tcap,\min_{j\in\mathcal A(H)}Y_{ej}(X,H)\},
\quad \Tcap=4\ \mathrm{s}.
\label{eq:scene-pet}
\end{align}
The infimum of an empty set is infinity. The cap therefore groups both sufficiently separated encounters and futures with no shared occupancy within the window. Minimum PET thus has point masses at zero, when two footprints occupy shared space at the same time, and at the cap. The compatible interval of Eq.~\eqref{eq:rank} scores outcomes at both consistently. Taking the minimum over SVs lets the joint future determine which encounter dominates scene risk.

History contains 13 observed states spanning 0.96 seconds, and the output contains 175 states at 25 Hz, including the common initial state. The number of vehicles varies by clip, and every vehicle selected from the history appears in the generated future. Clips enter the training population only when every selected vehicle is observed throughout the future window. Model inputs use the observed history and static scene information. For the capped PET implementation, endpoint requests use $q_H(0)=\Tcap$ and $q_H(1)=0$, and experiments use interior requests. Applying the percentile formulation to another scalar surrogate requires reference data and a generation mechanism suited to that measurement.

\begin{figure*}[!t]
\centering
\includegraphics[width=.7\textwidth,height=.82\textheight,keepaspectratio]{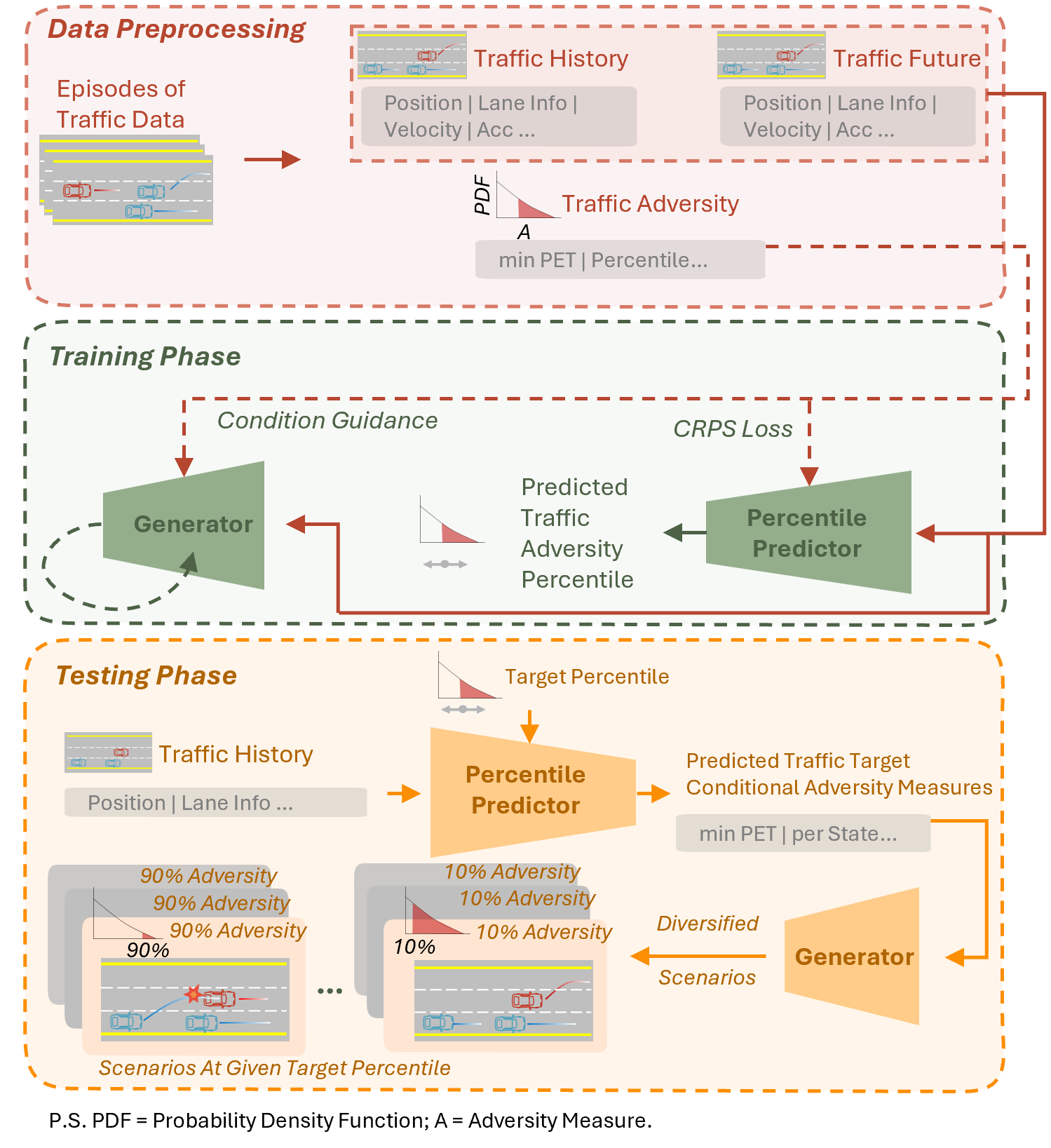}
\caption{Three-layer workflow: data preprocessing, training and testing. The percentile predictor denotes the reference risk distribution, trained by CRPS on observed scene-risk measurements. Leave-recording-out references supply percentile labels for generator training, and the calibrated reference fitted on all training recordings translates a test-time request into a physical risk target. The request enters the joint generator directly and guides its sampling path, together with road and background-separation constraints.}
\label{fig:main}
\end{figure*}

\section{An executable reference-to-generation framework}
\label{sec:method}
The request of Section~\ref{sec:problem} does not prescribe how its reference is estimated or how its target is realized. Any calibrated conditional distribution can serve as the reference, and any generator that can be steered toward a physical target can realize the request. The implementation below is one such design for multi-vehicle highway scenes, and it is not necessarily the best one.

The reference and generator turn a relative request into a joint traffic future in two stages. First, the reference risk distribution estimates the distribution of natural scene risk for the observed history, giving each percentile a physical risk target. Then the generator uses the request, its target and the reference distribution to produce the future. Fig.~\ref{fig:main} shows how these stages connect. We refer to the complete system, which is conditioned on the physical risk target and guided toward it during sampling, as our method. Reference fitting and calibration precede generator training. Percentile labels for training come from leave-recording-out references, each fitted without the recordings whose clips it labels. At generation time, the calibrated reference fitted on all training recordings translates the user's $p$ into its target.

\subsection{Learning a distribution from historical interactions}
The reference must relate past interactions to a distribution of possible scene-risk outcomes. Each natural clip contributes one realized future. We use these observations to train a shared conditional distribution model with CRPS, allowing statistical information to be shared across histories. The model first encodes vehicle motion and ego-relative interactions, then reads those interactions separately for different parts of the risk distribution.

Separate actor-shared temporal Transformers encode absolute vehicle histories and ego-relative SV histories. The absolute stream captures vehicle motion, and the relative stream captures how each SV has moved with respect to the ego. Dimensions are fused with actor embeddings, while pooled road features and vehicle count form a scene context $c_H$. With ego representation $h_e$, SV representation $h_j$, and relative-history representation $\Delta h_{ej}$, we form relation tokens
\[
u_j=\phi_{\rm pair}([h_e,h_j,\Delta h_{ej},c_H]).
\]
An ego-context token supplies a scene anchor. Padding is masked throughout. Temporal summary-token encoding follows the general approach used in motion-history representation \citep{zhou2022hivt}.

Different risk ranges can draw on different historical relations. We divide $(0,\Tcap)$ into $B=64$ bins and represent the two endpoints separately. Each component uses its PET range to query the relation tokens:
\begin{align}
q_b&=\phi_q([h_e,c_H,\phi_{\rm range}(b)]),\\
v_b&=\operatorname{Attention}(q_b,\{u_j\}\cup\{u_e\}),\\
\ell_b&=\phi_{\rm out}([h_e,q_b,v_b]).
\label{eq:range-attention}
\end{align}
This gives the small-gap and capped components their own views of the same historical interactions. Each query's range is fixed by its output distribution component. The readout jointly uses the relation set to predict the distribution of the scene minimum directly.

A joint softmax converts the 66 logits into masses $(m_0,m_1,\ldots,m_B,m_\tau)$. Writing $U_b$ for a uniform-bin CDF gives
\begin{equation}
\begin{aligned}
F_{\psi,H}(y)={}&m_0\ind\{y\ge0\}
+\sum_{b=1}^{B}m_bU_b(y)\\
&+m_\tau\ind\{y\ge\Tcap\}.
\end{aligned}
\label{eq:mixed}
\end{equation}
The mixed CDF represents continuous PET variation together with the zero and capped outcomes, and is normalized and monotone by construction. CDF, left-limit and inverse queries all follow from this probability representation.

CRPS trains the full distribution from the physical outcome observed in each clip. For observed PET values $y_i$, training minimizes the normalized CRPS,
\begin{equation}
\mathcal L_A=\frac{1}{n\Tcap}\sum_i
\int_0^\Tcap
\left[F_{\psi,H_i}(y)-\ind\{y_i\le y\}\right]^2dy .
\label{eq:crps}
\end{equation}
At each threshold, the prediction is compared with the observed threshold event, and integration trains the distribution across its range \citep{gneiting2007strictly}. The observed physical outcome therefore supplies distributional supervision, while percentile labels are derived from the fitted reference later. Integration is analytic on the CDF's linear pieces.

After fitting, a separate calibration set adjusts the reference's probability scale. A monotone probability-axis warp produces
\begin{equation}
\G_H(y)=h_{N(H)}(F_{\psi,H}(y)).
\label{eq:warp}
\end{equation}
The count-conditioned warp follows distribution recalibration principles \citep{kuleshov2018accurate,song2019distribution}. The composed CDF keeps endpoint masses and any additional physical breakpoints, so its probability, left-limit and target queries remain consistent. Section~\ref{sec:experiments} evaluates the predictive quality of this reference separately from request realization.

\subsection{Direct percentile-conditioned joint diffusion}
The generator learns to turn the reference-based request into coordinated vehicle motion. It receives $H$, the requested $p$, and noise $z$. Its risk condition combines the scalar request, the physical risk target $q_H(p)$, and a descriptor of $\G_H$. The target specifies the physical outcome to realize, while the CDF descriptor locates it within the history-specific scale. A separate risk-conditioning pathway modulates actor tokens during denoising.

We generate the future through a compact motion representation that preserves the observed initial state. Each vehicle's future acceleration is represented by eight cosine coefficients per coordinate. Collect these coefficients in $c$ and denote the analytic decoder by $D_H$. For coordinate $a$,
\begin{equation}
x_{i,a}(t)=x_{i,a}(0)+v_{i,a}(0)t+
\sum_{k=0}^{7}c_{i,k,a}B_k(t).
\label{eq:decoder}
\end{equation}
The integrated basis fixes initial position and velocity exactly. A lightweight Transformer denoises the joint coefficient tensor, using masked actor attention to couple the vehicles of scenes of any size. Its actor tokens combine observed history, dimensions, ego role, road context, diffusion time and noisy coefficients. The generator uses a multilayer perceptron (MLP) to map history into these tokens, separately from the temporal reference encoders.

For normalized clean coefficients $c_0$, diffusion time $t$ and standard Gaussian noise $\epsilon$, define
\[
c_t=\alpha_t c_0+\sigma_t\epsilon,\quad
v_t=\alpha_t\epsilon-\sigma_tc_0,
\quad \alpha_t^2+\sigma_t^2=1.
\]
Training uses a masked $v$-prediction mean squared error (MSE) \citep{ho2020denoising,salimans2022progressive}. Risk-condition dropout also trains a null-risk branch that keeps the historical and static scene inputs. Classifier-free guidance combines the two predictions,
\begin{equation}
v_{\rm CFG}=v_{\rm null}+s(v_{\rm cond}-v_{\rm null}),
\qquad s=2.5,
\label{eq:cfg}
\end{equation}
during 50 deterministic steps of denoising diffusion implicit models (DDIM) \citep{ho2022classifier,song2021denoising}.

Natural futures supply the trajectory targets, and the leave-recording-out references assign their percentile conditions. A final risk-aware training stage keeps the natural-data objective and additionally trains futures sampled for auxiliary requests toward their percentile and physical targets. In both training and sampling, physical targets follow the definition in Eq.~\eqref{eq:percentile-target}.

\subsection{Realizing the physical risk target within one sampling path}
Direct conditioning learns how the joint future should respond to $p$, and sampling-time risk guidance brings its physical outcome toward $q_H(p)$. During the final 15 DDIM steps, we decode the clean-coefficient estimate and evaluate minimum ego--SV PET. The encounter attaining this minimum can change as trajectories move. We therefore use a local active-geometry derivative to propose a direction, then evaluate a feasible one-sided perturbation with the complete PET query.

Let $c$ be the current feasible estimate, $u$ a projected probe displacement, $Y_c=Y(D_H(c),H)$ and $\Delta Y=Y(D_H(c+u),H)-Y_c$. When the measured difference has the required sign, the proposed update is
\begin{equation}
\Delta c=-\eta\frac{Y_c-q_H(p)}{\Delta Y}u,
\qquad \eta=0.8.
\label{eq:guide}
\end{equation}
A fixed position--velocity--acceleration metric preconditions directions. Each risk increment changes positions by at most 0.5 meters, with at most six internal updates per enabled step. Where small displacements leave the PET unchanged, as at the cap or at zero, explicit spatial signals that bring two vehicles together or apart supply the direction.

Road and background projections constrain the geometry of the sampled scene. Road constraints keep footprints within the outer road boundaries, permit internal lane crossings, and tolerate any boundary exceedance already present in the observed initial state. Background separation is activated during the final five denoising steps. For each background pair, lateral-overlap windows determine a longitudinal ordering and constraints of the form
\begin{equation}
s_{ij}\{x_j(t)-x_i(t)\}
\ge(\ell_i+\ell_j)/2+\delta,\qquad \delta=0.1\ \mathrm{m},
\label{eq:bg}
\end{equation}
where $s_{ij}\in\{-1,1\}$ records that ordering. Small coefficient-space quadratic programs enforce the active constraints. Lateral separation can allow a different ordering in a later window.

Risk guidance and scene constraints operate within one noise path ($K=1$), before each DDIM transition in which they are active and before the final output. Each request therefore yields one generated future.

\section{Empirical evaluation on natural traffic}
\label{sec:experiments}
The experiments trace a request from its definition to its realization. Section~\ref{sec:final-reference} establishes the need for a contextual scale and evaluates the learned reference that provides it. Sections~\ref{sec:final-generation} and~\ref{sec:p-path-ablation} measure how closely generated futures occupy the requested position and identify the part of our method responsible. Section~\ref{sec:external-no-p} turns to natural priors that receive no request, and Section~\ref{sec:qualitative-scenes} illustrates the generated interactions. Section~\ref{sec:kinematic-realism} checks that precise control leaves the motion plausible. Section~\ref{sec:ttc-control} carries the interface over to time to collision (TTC) in car following, and Section~\ref{sec:planner-test} uses the generated scenarios to test a rule-based planner.

\subsection{Data, requests and evaluation}
All observations come from natural highD trajectories \citep{krajewski2018highd}, split by recording into training, validation and test groups. The reference and generator are fitted on 9,913 complete training clips, with separate early-stopping and calibration sets of 538 and 504 clips. The primary evaluation set of 11,649 clips comes from the 16 test recordings and is used unless stated otherwise. An additional set of 9,277 clips from the 14 validation recordings and a single-recording set of 603 clips from a recording outside the split provide further tests. Earlier versions of the models were also evaluated on the test and validation recordings, and the models reported here were fixed before the final evaluation, with no evaluation recording used for fitting, selection or calibration. At generation time, models observe only the history and static scene information.

Generation uses 96 primary-set histories, balanced across recordings, with 32 in each vehicle-count group $N=3$--5, 6--8 and at least 9. Each history receives the requests $p=0.1,0.3,0.5,0.7,0.9$ with the same three noise draws, which gives 1,440 requests. Each request yields one future ($K=1$), and every future is scored, so failures count against the method. The additional and single-recording sets supply 96 and 70 histories with the same requests and noise draws.

Realization is measured by three quantities. P-MAE is the mean percentile error. The fine-control rate $\Fine$ is the percentage of requests realized within 0.05 of the requested percentile. PET-target MAE is the mean physical error with respect to the target $q_H(p)$. Footprint overlap and strict road compliance, under which any exceedance of the outer road boundary counts as a violation, describe the generated scenes. Because minimum PET has point masses at zero and at the four-second cap, every compared generator is scored with the interval-compatible criterion of Eqs.~\eqref{eq:rank}--\eqref{eq:risk-mae-goal}.

\subsection{A contextual scale and its learned reference}
\label{sec:final-reference-test}
\label{sec:final-reference}
\begin{figure}[t]
\centering
\includegraphics[width=.6\linewidth]{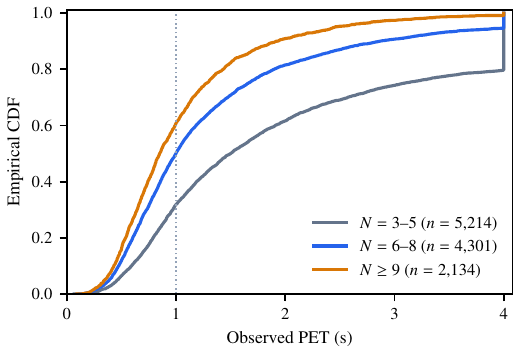}
\caption{Empirical PET CDFs of 11,649 natural clips for histories with 3--5, 6--8 and at least nine vehicles, with group sizes in the legend. The dotted line marks PET $=1$ second. Group-level outcome variation motivates the contextual scale.}
\label{fig:reference-diagnostics}
\end{figure}

A percentile request presupposes that natural risk shifts with traffic context. Fig.~\ref{fig:reference-diagnostics} confirms the shift. At a PET of one second, the empirical CDF rises from 0.319 for histories with 3--5 vehicles to 0.609 for histories with at least nine. The same one-second PET lies in the lower third of natural outcomes in sparse traffic and above the median in dense traffic. A fixed physical threshold would thus signal different adversity in different contexts, whereas a percentile keeps its meaning across them. The reference conditions on the full observed history, and the count groups only make this dependence visible.

\begin{table*}[tp]
\centering
\scriptsize
\setlength{\tabcolsep}{3pt}
\caption{Reference evaluation on 11,649 natural clips. Lower is better. CRPS is in seconds and is given overall and for histories with $N=3$--5, 6--8 and at least nine vehicles (5,214, 4,301 and 2,134 clips). PIT$_{\rm tail}$: largest deviation of the expected CDF of the randomized probability integral transform (PIT) from the diagonal at levels 0.05--0.30, where the targets of the critical requests $p\ge0.7$ lie. Cap Brier: Brier score for PET reaching the four-second cap. Classical estimators use pooled history features without calibration. The static-query reference is our dual-stream architecture with range-independent queries. Bold marks the best value in each column.}
\label{tab:canonical-reference}
\begin{tabular}{@{}lrrrrrr@{}}
\toprule
 & \multicolumn{4}{c}{CRPS $\downarrow$} & & \\
\cmidrule(lr){2-5}
Reference & All & $N=3$--5 & $N=6$--8 & $N\ge9$ & PIT$_{\rm tail}\downarrow$ & Cap Brier $\downarrow$\\
\midrule
Unconditional ECDF & 0.60387 & 0.75103 & 0.51421 & 0.42501 & 0.03461 & 0.09969\\
kNN conditional CDF \citep{stone1977consistent} & 0.40331 & 0.46997 & 0.38083 & 0.28574 & 0.02059 & 0.04256\\
Quantile regression forest \citep{meinshausen2006quantile} & 0.33817 & 0.37500 & 0.33036 & 0.26391 & \textbf{0.00402} & 0.02735\\
Static-query reference & 0.11160 & 0.15139 & 0.08866 & \textbf{0.06061} & 0.04318 & 0.02003\\
Range-query reference (ours) & \textbf{0.10550} & \textbf{0.14041} & \textbf{0.08504} & 0.06144 & 0.01051 & \textbf{0.01722}\\
\bottomrule
\end{tabular}
\end{table*}
\begin{table*}[tp]
\centering
\scriptsize
\setlength{\tabcolsep}{3pt}
\caption{Percentile-request realization and scene geometry of the compared generators for 1,440 requests (96 histories, five percentiles, three noise draws). Methods without a percentile input are scored against all five requests of each history. TrafficGen gives one output per history by top-1 decoding, and CTG++ and STRIVE give 15. RADE receives the physical risk target of each request, converted to its risk level. Fine: fine-control rate, the percentage of requests realized within 0.05 of the requested percentile (interval error at most 0.05). P-MAE: mean percentile error. PET-MAE: mean absolute PET-target error (s). BG, Ego and Road: percentages of requests with background--background overlap, ego-related overlap and a strict road violation. Bold marks the best value in each column.}
\label{tab:canonical-control}
\begin{tabular}{@{}lrrrrrr@{}}
\toprule
Method & Fine (\%) $\uparrow$ & P-MAE $\downarrow$ & PET-MAE $\downarrow$ & BG (\%) $\downarrow$ & Ego (\%) $\downarrow$ & Road (\%) $\downarrow$\\
\midrule
\multicolumn{7}{l}{\emph{External methods without a percentile input}}\\
TrafficGen motion \citep{feng2022trafficgen} & 16.04 & 0.28665 & 0.14083 & 2.08 & 1.04 & 52.08\\
CTG++ unguided prior \citep{ctgpp2023} & 14.00 & 0.33789 & 0.19553 & 0.21 & \textbf{0.00} & 21.25\\
STRIVE traffic prior \citep{rempe2022strive} & 13.06 & 0.31757 & 0.26149 & 6.67 & 1.67 & 75.49\\
\midrule
\multicolumn{7}{l}{\emph{External method with a physical risk input}}\\
RADE \citep{wang2025rade} & 14.31 & 0.31555 & 0.20461 & 7.50 & 3.89 & 34.17\\
\midrule
\multicolumn{7}{l}{\emph{With a percentile input}}\\
P-CVAE \citep{sohn2015learning} & 16.25 & 0.32465 & 0.21517 & 11.11 & 2.08 & 57.50\\
Standard P-diffusion \citep{ho2020denoising} & 17.01 & 0.29226 & 0.15928 & 2.64 & 1.18 & 21.53\\
Our method & \textbf{98.75} & \textbf{0.00673} & \textbf{0.00991} & \textbf{0.00} & \textbf{0.00} & \textbf{5.76}\\
\bottomrule
\end{tabular}
\end{table*}
Every request is defined through the reference, so its accuracy underlies all later results. Table~\ref{tab:canonical-reference} compares the learned reference with three classical estimators. The unconditional empirical CDF (ECDF) assigns every history the distribution of all training outcomes. The kNN conditional CDF uses the outcomes of the 32 training histories nearest to the query history \citep{stone1977consistent}, and the quantile regression forest weights training outcomes by how often they share a leaf with the query history across 100 regression trees \citep{meinshausen2006quantile}. Both operate on a fixed vector of pooled history features. The learned references lower CRPS by a factor of three to six, from 0.33817 seconds for the best classical estimator to 0.10550 for the range-query reference. In short, the relational references are far sharper than estimators on pooled history features.

How the relations are read out also matters. Within the same dual-stream architecture, range-dependent queries reduce CRPS by 5.47\%, from 0.11160 seconds for the static-query reference to 0.10550, and a paired bootstrap over recordings gives a 95\% interval of $[-0.00894,-0.00314]$ seconds for the difference. They also calibrate the critical tail, where the targets of high requests lie, and lower PIT$_{\rm tail}$, the tail error of the probability integral transform (PIT), from 0.04318 to 0.01051. The quantile regression forest is calibrated even more closely in the tail, but its CRPS is three times higher, so its targets would follow the individual history far less. The largest PIT deviation of the range-query reference, 0.07915, lies on the benign side of the distribution, at the levels of low requests (Appendix~\ref{app:context-diagnostics}). The range-query reference supplies all targets and scores below. It has the lowest CRPS on the development set (Appendix~\ref{app:context-diagnostics}), and it is one implementation of the reference that need not be the most accurate. Among the encoders compared in Appendix~\ref{app:context-diagnostics}, one that reads only absolute histories is slightly more accurate.

\subsection{Realizing the requested percentile}
\label{sec:final-generation}
\label{sec:supplemental-comparison}
The main experiment measures whether a generated future lands at the requested position. Our method meets 1,422 of the 1,440 requests (98.75\%) within the 0.05 tolerance, with P-MAE 0.00673 and PET-target MAE 0.00991 seconds (Table~\ref{tab:canonical-control}). Requests with continuous targets, which demand fine physical control, are met in 98.65\% of cases, and the 111 requests whose target is a point mass are all matched exactly. The additional and single-recording sets confirm the result with 95.07\% and 99.24\%. In nearly every case, a percentile request yields a future at the requested position of its own history.

Two percentile-conditioned baselines show what learned conditioning achieves on its own. P-CVAE is a conditional variational autoencoder \citep{sohn2015learning} that decodes a joint future from the history, $p$ and a latent sample. Standard P-diffusion is a conditional diffusion model \citep{ho2020denoising} that receives $p$ as a conditioning input and samples with DDIM \citep{song2021denoising}. Both use the trajectory representation and leave-recording-out percentile labels of our method. They meet only 16.25\% and 17.01\% of the requests, close to generators that ignore the request (Section~\ref{sec:external-no-p}). Learning to condition on $p$ leaves the realized percentile largely uncontrolled.

Risk-conditioned generation is the most direct alternative to a percentile request. RADE conditions joint multi-agent diffusion on a risk level derived from scene PET \citep{wang2025rade}. Its code is not public, so we reproduce its conditioning on the Standard P-diffusion backbone. For each request, RADE receives the physical risk target $q_H(p)$, converted to its risk level. Its scenes become more severe as the risk level rises, but it meets only 14.31\% of the requests. A physical risk level thus shifts the overall severity of the scenes, yet it controls the position of each request within its own context no better than learned conditioning on $p$.

\begin{figure*}[tp]
\centering
\includegraphics[width=\linewidth]{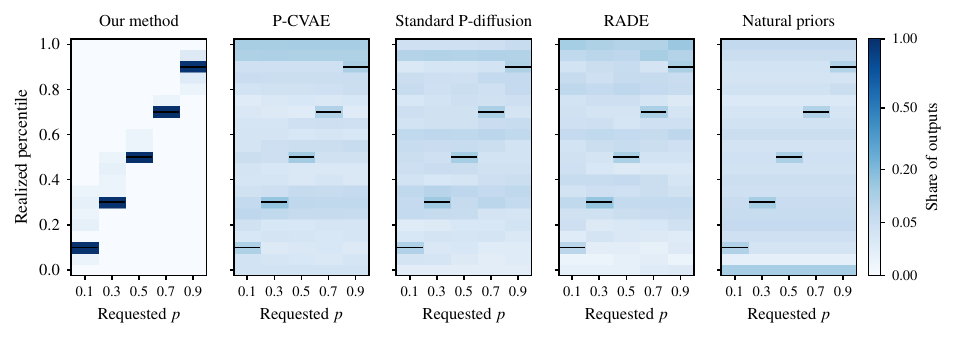}
\caption{Where the outputs land. Each column shows, for one requested percentile, the share of outputs whose realized percentile falls in each 0.05-wide bin, on a square-root color scale. An output at a probability mass is placed at the point of its percentile interval nearest the request, as in the interval criterion. Black marks show the request. Natural priors receive no request, so each of their outputs is placed against all five requests.}
\label{fig:calibration-heatmap}
\end{figure*}

\begin{figure*}[tp]
\centering
\includegraphics[width=\linewidth]{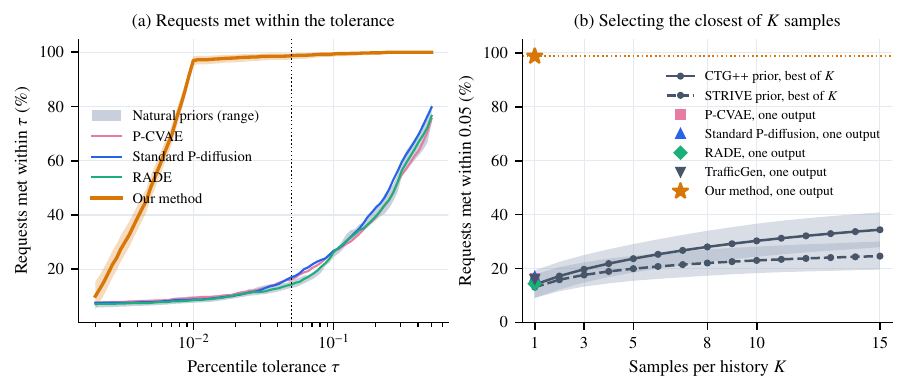}
\caption{Precision across tolerances and selection among samples. (a) Share of the 1,440 requests realized within a percentile tolerance $\tau$. The band around our method is its 95\% cluster-bootstrap interval over the 96 histories, the gray band spans the three natural priors, and the dotted line marks the 0.05 tolerance used throughout. (b) Share of requests realized within 0.05 when the sample whose realized percentile is closest to the request is chosen among $K$ samples of the same history, for the two priors with 15 samples per history. Values are exact averages over all subsets of $K$ samples, with 95\% cluster-bootstrap bands. The other generators give one output per request, except TrafficGen with one output per history, and appear at $K=1$.}
\label{fig:tolerance-cost}
\end{figure*}

Fig.~\ref{fig:calibration-heatmap} shows where the outputs land. Those of our method concentrate at the requested percentile, while the outputs of the compared generators spread over the whole scale at every request, as do those of natural priors that receive no request. Sampling variation cannot account for this gap, since the 95\% cluster-bootstrap interval of our Fine starts at 97.43\% and those of all compared generators end at or below 21.81\%. The gap holds at every tolerance (Fig.~\ref{fig:tolerance-cost}(a)), and drawing more samples does not close it (Fig.~\ref{fig:tolerance-cost}(b)). Even when the sample whose percentile is closest to the request is chosen among 15 samples of the same history, the CTG++ prior meets 34.4\% of the requests and the STRIVE prior 24.6\%. The samples of a prior cover only part of the distribution of their history, so for most requests none of them comes within the tolerance. A requested percentile has to enter generation to be reached reliably.

\subsection{Which P pathway realizes the request?}
\label{sec:p-path-ablation}
\begin{figure*}[tp]
\centering
\includegraphics[width=\linewidth]{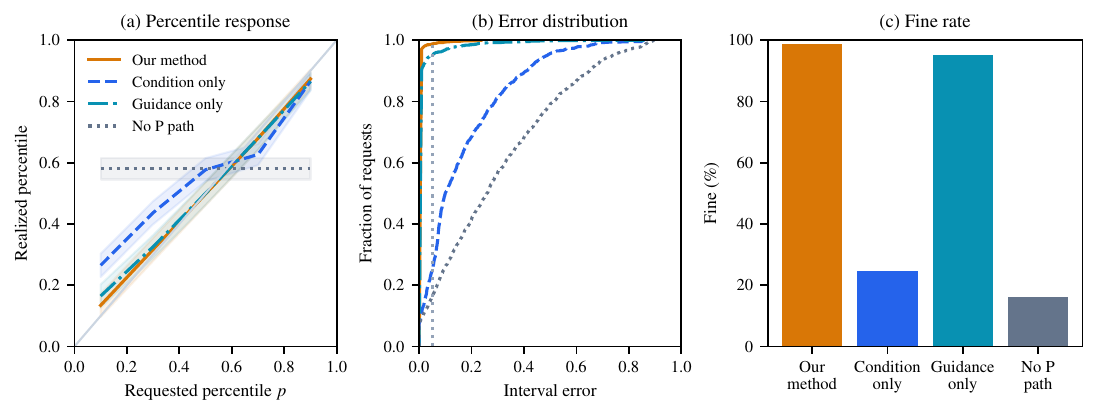}
\caption{P-path ablation over the 1,440 requests. (a) Realized percentile against requested percentile for Our method, Condition only, Guidance only and No P path. Shaded bands span the mean lower and upper ends of the compatible percentile interval, and lines mark the center of each band. (b) Cumulative distribution of the interval error, with a dotted line at the 0.05 tolerance. (c) Fine, the percentage of requests realized within the 0.05 tolerance. Together, the panels show the dominant precision gain from guidance and the further improvement from learned conditioning.}
\label{fig:p-path-control}
\end{figure*}

Our method uses the request in two places: a learned risk-conditioning pathway in the denoiser and sampling-time risk guidance. To identify which of them carries the request, the ablation switches them off separately and jointly, with the same trained weights, reference, histories and noise draws (Table~\ref{tab:p-path-ablation} and Fig.~\ref{fig:p-path-control}). Without the conditioning pathway, the trained null-risk branch still receives the history and static scene context. Road and background constraints remain active in every configuration. Without either pathway (No P path), the output does not depend on $p$, so each future is scored against all five requests.

\begin{table*}[tp]
\centering
\footnotesize
\setlength{\tabcolsep}{3pt}
\caption{P-path ablation of our method on the same 1,440 requests. All configurations share the trained weights, reference, histories and noise draws, and road and background constraints are active in every row. Condition only keeps the learned risk-conditioning pathway, Guidance only keeps sampling-time risk guidance, and No P path removes both. Without either pathway, the output does not depend on $p$, and one future per history and noise draw is scored against all five requests. Metrics as in Table~\ref{tab:canonical-control}. Bold marks the best value in each column in which the configurations differ.}
\label{tab:p-path-ablation}
\begin{tabular}{@{}lrrrrrr@{}}
\toprule
Configuration & Fine (\%) $\uparrow$ & P-MAE $\downarrow$ & PET-MAE $\downarrow$ & BG (\%) $\downarrow$ & Ego (\%) $\downarrow$ & Road (\%) $\downarrow$\\
\midrule
Our method & \textbf{98.75} & \textbf{0.00673} & \textbf{0.00991} & 0.00 & \textbf{0.00} & \textbf{5.76}\\
Condition only & 24.79 & 0.16554 & 0.11612 & 0.00 & 0.49 & 6.11\\
Guidance only & 95.07 & 0.01469 & 0.01688 & 0.00 & \textbf{0.00} & 6.18\\
No P path & 16.18 & 0.29792 & 0.16246 & 0.00 & 0.69 & 6.25\\
\bottomrule
\end{tabular}
\end{table*}
Guidance provides most of the precision. On its own it reaches 95.07\% of the requests, whereas conditioning alone gives 24.79\% and removing both leaves 16.18\% (Fig.~\ref{fig:p-path-control}(c)). Adding conditioning to guidance raises Fine by a further 3.68 percentage points, to 98.75\%, and more than halves P-MAE, from 0.01469 to 0.00673 (Fig.~\ref{fig:p-path-control}(a) and (b)). The two pathways are complementary: the learned pathway supplies a useful request-dependent starting point, and explicit target realization during sampling makes it precise. What decides realization is sampling toward the physical target that the reference derives from the request. Conditioning on $p$ realizes few requests, even when the target is part of the condition as in Condition only, and so do Standard P-diffusion and P-CVAE (Section~\ref{sec:final-generation}). Guidance toward the target realizes most requests even without the learned conditioning pathway. Percentile control thus rests on defining the request through the reference of its history and steering the sample to the resulting target, and learned conditioning adds precision on top of it.

\begin{figure*}[tp]
\centering
\includegraphics[width=\linewidth]{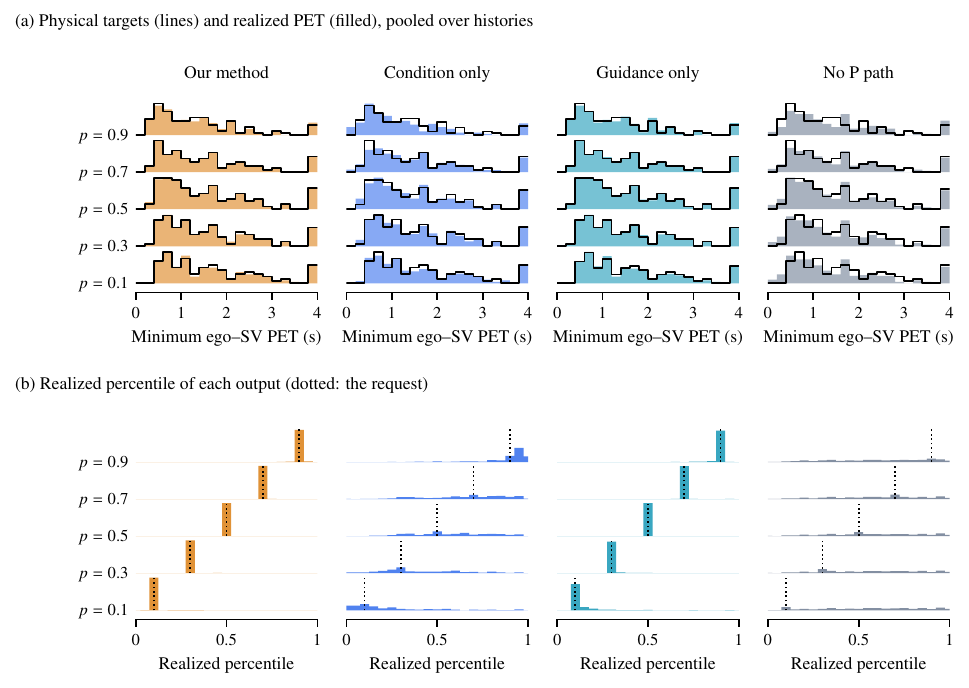}
\caption{Physical targets and realized percentiles of the four configurations of Table~\ref{tab:p-path-ablation} over the 1,440 requests. (a) At each request, the distribution of the physical targets $q_H(p)$ (lines) and of the realized minimum ego--SV PET (filled), pooled over the 96 histories and three noise draws. The last bin collects outcomes at the 4~s cap. (b) The realized percentile of each output at each request in 0.05-wide bins, placed at the point of its percentile interval nearest the request as in Fig.~\ref{fig:calibration-heatmap}. Dotted lines mark the request. No P path generates one future per history and noise draw and scores it against all five requests.}
\label{fig:p-path-ridgelines}
\end{figure*}

Fig.~\ref{fig:p-path-ridgelines} contrasts two views of the same outputs. Pooled over histories, the realized PET of every configuration stays close to the distribution of the physical targets at each request (Fig.~\ref{fig:p-path-ridgelines}(a)), because the targets of different histories spread over the whole range. The realized percentile of each output separates the configurations (Fig.~\ref{fig:p-path-ridgelines}(b)). Our method and guidance only place almost every output at its request, whereas conditioning alone and No P path spread their outputs over the whole scale, although their median PET errors at each request range only from 0.03 to 0.11~s. A pooled comparison of physical outcomes therefore cannot show whether a generator controls the percentile.

\subsection{What external priors generate without a request}
\label{sec:external-no-p}
Published scenario generators learn futures from natural data and take no percentile input. Scoring their outputs against the same requests shows how far such priors are from percentile control. TrafficGen \citep{feng2022trafficgen} places vehicles and then generates their motion, and we use its motion stage from the observed initial scene with top-1 decoding. CTG++ \citep{ctgpp2023} is a scene-level diffusion model whose sampling can follow rules given in natural language, and we sample its unguided prior. STRIVE \citep{rempe2022strive} optimizes the latent codes of a graph-based conditional VAE traffic prior against a planner to create accident-prone scenarios, and we sample the prior without this optimization. All three are trained on the same natural training clips and receive the history and static context of the same 96 histories. TrafficGen gives one output per history, and CTG++ and STRIVE give 15. As with No P path, each output is scored against all five requests of its history.

The priors place 13--16\% of output--request pairs within the tolerance (Table~\ref{tab:canonical-control}), about the same level as No P path and the percentile-conditioned baselines. A natural prior covers a range of adversity, yet the position of any single output is left to chance. Hitting a requested position takes a generator that uses its target.

\subsection{Generated multi-vehicle behavior}
\label{sec:qualitative-scenes}
\begin{figure*}[tp]
\centering

\includegraphics[width=.93\linewidth,height=.78\textheight,keepaspectratio]{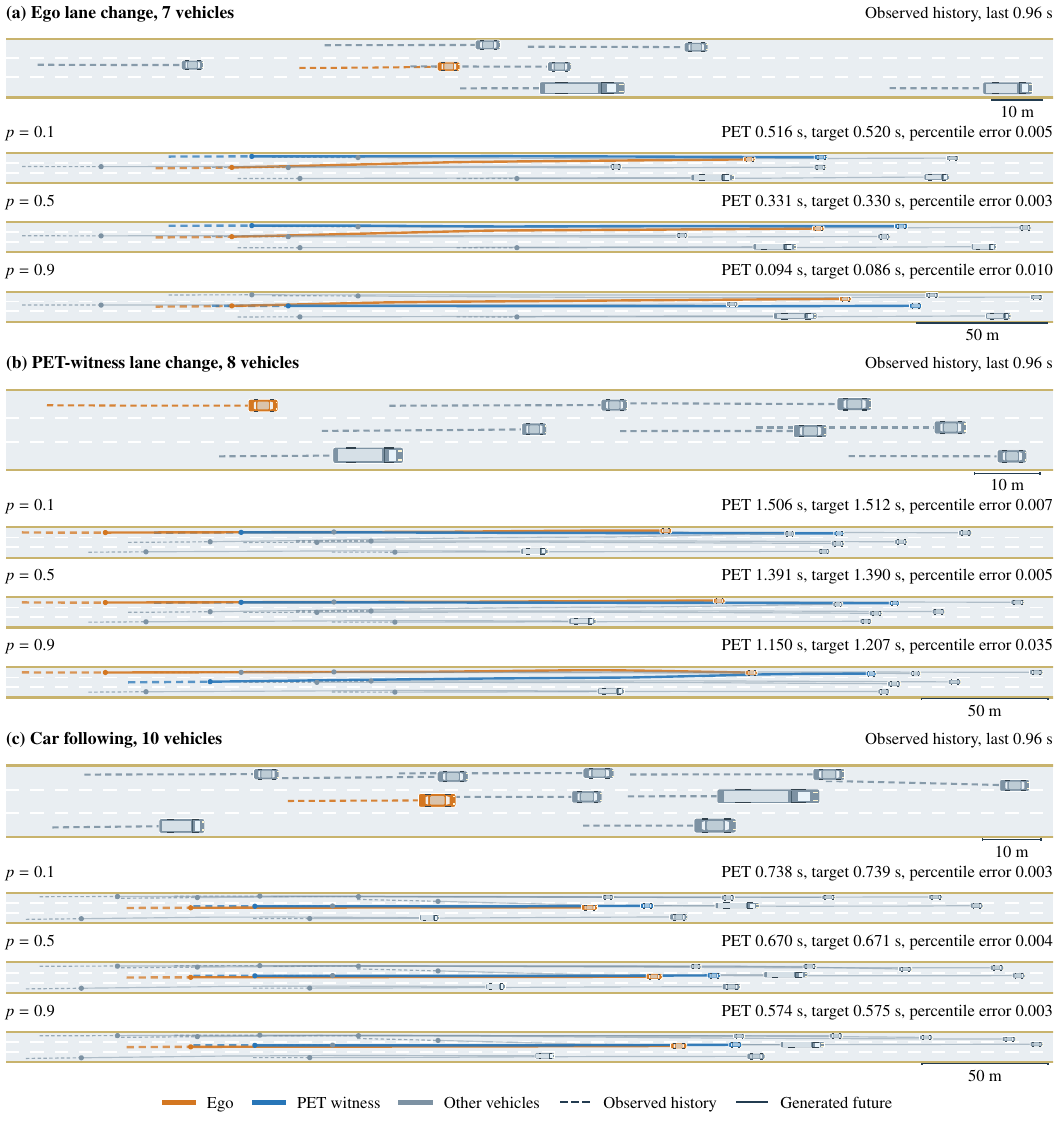}

\caption{Lane changes and car following generated by our method. Each block shows the observed history (the last 0.96 seconds, at a larger scale) and the futures generated from it with one noise draw at $p=0.1,0.5,0.9$, and each row reports the realized PET, its target and the percentile error. (a) A seven-vehicle history in which the ego changes lanes at all three requests. (b) An eight-vehicle history in which the PET witness at $p=0.9$ changes lanes. (c) A ten-vehicle history in which the ego follows its PET witness, the lead vehicle in the same lane, at all three requests. Orange marks the ego, blue the PET witness of each future and gray the other vehicles. Observed histories are dashed and generated futures solid. Vehicle bodies are drawn at the last observed state in the history views and at 6.96 seconds in the future rows, and dots mark positions at the start of the future. All vehicles are shown with their measured dimensions, and road views use equal $x$ and $y$ scales. The examples were selected to illustrate these interactions.}
\label{fig:qualitative-scenes}
\end{figure*}

The same requested position can arise from quite different ego--SV interactions. We call the SV that attains the minimum PET of a future its PET witness. Fig.~\ref{fig:qualitative-scenes} presents three histories at $p=0.1,0.5,0.9$. In Fig.~\ref{fig:qualitative-scenes}(a), the ego changes lanes at every request, and a different SV becomes its witness at $p=0.9$. In Fig.~\ref{fig:qualitative-scenes}(b), the witness itself changes lanes at $p=0.9$. In Fig.~\ref{fig:qualitative-scenes}(c), the ego follows its witness in the same lane, and as $p$ increases both vehicles travel faster at similar spacing, which shortens the time gap. Each realized PET in this example lies within about one millisecond of its target. Appendix~\ref{app:generated-lane-change-selection} describes how the examples were selected.

\begin{figure*}[tp]
\centering

\includegraphics[width=\linewidth,height=.78\textheight,keepaspectratio]{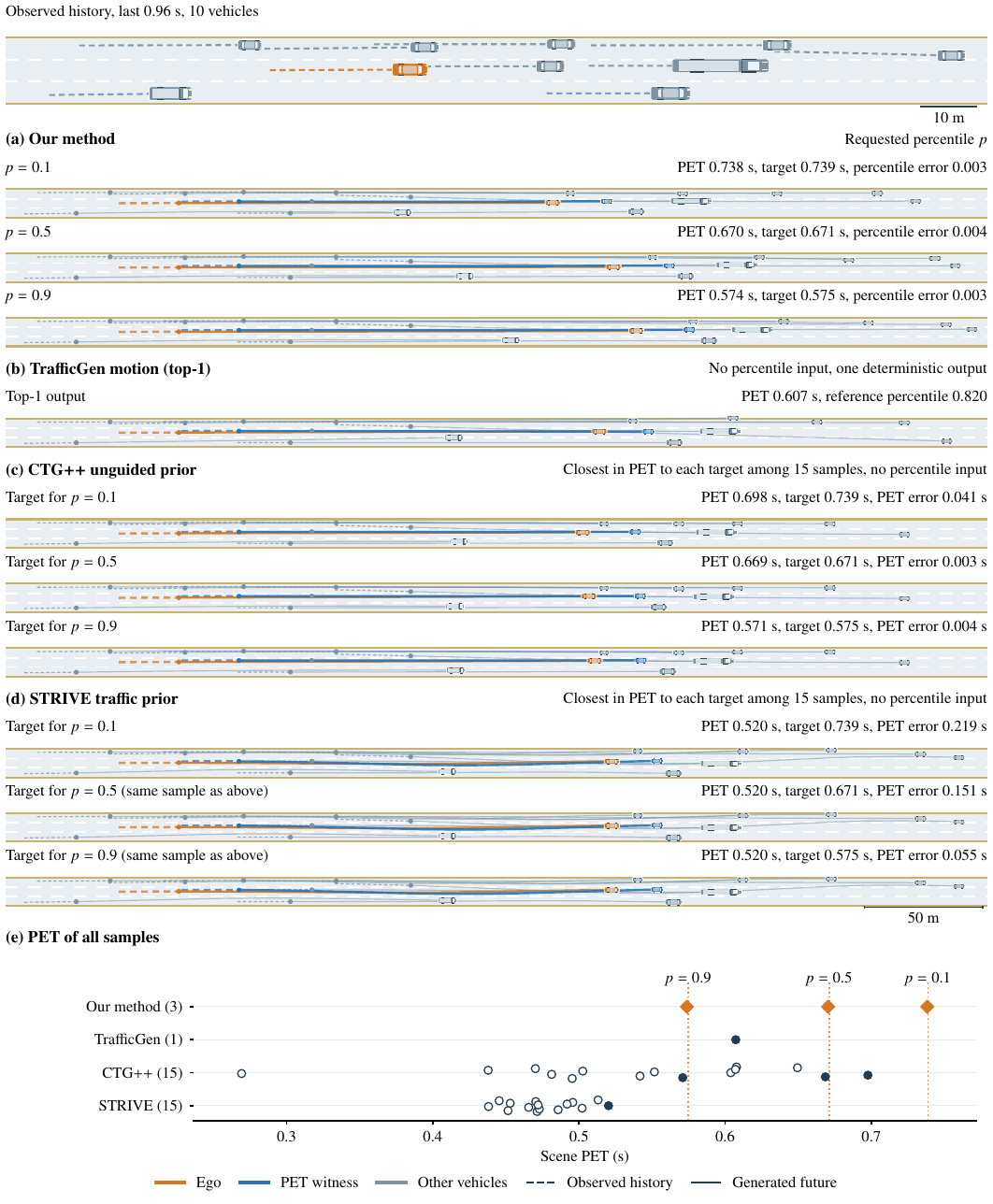}

\caption{Direct percentile conditioning and PET-based selection from external priors on the car-following history of Fig.~\ref{fig:qualitative-scenes}(c). (a) Our method at $p=0.1,0.5,0.9$, the futures shown in Fig.~\ref{fig:qualitative-scenes}(c). (b) TrafficGen motion: top-1 output, generated without a percentile input. (c) CTG++ unguided prior and (d) STRIVE traffic prior: for each target, the sample whose PET is closest to that target among the prior's 15 samples, selected after generation for display. A sample that is closest to several targets appears in each of their rows. (e) PET of all samples, with filled markers for the displayed samples and dotted lines at the targets. The top panel shows the observed history at a larger scale. The other road views share one extent, and all road views use equal $x$ and $y$ scales. Vehicle colors follow Fig.~\ref{fig:qualitative-scenes}.}
\label{fig:external-vehicle-scenes}
\end{figure*}

Footprint overlap is rare in the generated scenes. None of the 3,930 futures generated by our method across the three evaluation sets has background--background overlap, and one has ego-related contact. Most strict road violations are inherited from the observed initial scene, 75 of the 83 on the primary set.

Could selecting from a prior replace direct conditioning? Fig.~\ref{fig:external-vehicle-scenes} examines this on the car-following history of Fig.~\ref{fig:qualitative-scenes}(c), where each target picks the external sample whose PET is closest to it. This selection even uses outcomes observed after generation. The output of our method is nonetheless closer to every target than any TrafficGen, CTG++ or STRIVE sample (Fig.~\ref{fig:external-vehicle-scenes}(e)). The aggregate results in Table~\ref{tab:canonical-control} use all samples, and Fig.~\ref{fig:tolerance-cost}(b) measures such selection over all histories.

\subsection{Kinematic realism of generated futures}
\label{sec:kinematic-realism}
\begin{table}[tp]
\centering
\footnotesize
\setlength{\tabcolsep}{4pt}
\caption{Kinematic realism of generated futures on the 96 evaluation histories. Speed and longitudinal and lateral acceleration are finite differences of positions on a 0.2~s grid, pooled over all vehicles and times. W1 is the Wasserstein-1 distance to the observed futures of the same histories (m/s for speed, m/s$^2$ for acceleration). Harsh braking: longitudinal acceleration below $-4$~m/s$^2$. Implausible: acceleration magnitude above 8~m/s$^2$. Bold marks the smallest distance in each column.}
\label{tab:kinematic-realism}
\begin{tabular}{@{}lrrrrr@{}}
\toprule
Futures & W1 speed & W1 $a_x$ & W1 $a_y$ & Harsh braking (\%) & Implausible (\%)\\
\midrule
Observed futures & -- & -- & -- & 0.12 & 0.01\\
\midrule
TrafficGen motion & 0.323 & 0.445 & 0.097 & 0.00 & 0.00\\
CTG++ unguided prior & 1.129 & 0.375 & 0.130 & 0.00 & 0.00\\
STRIVE traffic prior & \textbf{0.105} & 0.583 & 0.185 & 0.00 & 0.00\\
RADE & 0.225 & 0.311 & \textbf{0.086} & 0.05 & 0.00\\
P-CVAE & 0.189 & 0.357 & 0.103 & 0.00 & 0.00\\
Standard P-diffusion & 0.242 & 0.413 & 0.101 & 0.00 & 0.00\\
Our method & 0.422 & \textbf{0.220} & 0.102 & 0.18 & 0.09\\
\midrule
\multicolumn{6}{l}{\emph{Our method by requested percentile}}\\
$p=0.1$ & 2.084 & 0.555 & 0.105 & 0.48 & 0.23\\
$p=0.3$ & 0.146 & 0.356 & 0.103 & 0.16 & 0.12\\
$p=0.5$ & 0.851 & 0.401 & 0.102 & 0.05 & 0.03\\
$p=0.7$ & 0.583 & 0.379 & 0.103 & 0.08 & 0.01\\
$p=0.9$ & 2.324 & 0.631 & 0.096 & 0.15 & 0.06\\
\bottomrule
\end{tabular}
\end{table}
Precise control should not come at the cost of implausible motion. Table~\ref{tab:kinematic-realism} compares the speed and acceleration of the generated futures with the observed futures of the same 96 histories, measuring every generator in the same way. Our method is the closest to the observed futures in longitudinal acceleration and matches the other generators in lateral acceleration. Harsh braking remains rare, at 0.18\% of the vehicle-time samples against 0.12\% in the observed futures, and Appendix~\ref{app:canonical-training} describes where the rare stronger accelerations occur.

The larger speed distance of our method reflects its intended response to the request. Low requests slow the traffic down and high requests speed it up, with the median speed moving from 25.2~m/s at $p=0.1$ to 29.8~m/s at $p=0.9$, so the departure from the observed speeds concentrates at the two extreme requests and is smallest at $p=0.3$. RADE and Standard P-diffusion keep nearly the same speed at every request, in line with their weak response to it. Extreme percentiles are rare outcomes of their histories by definition, so their kinematics are expected to depart from the pooled observed futures.

\subsection{Percentile requests on time to collision}
\label{sec:ttc-control}
\begin{table}[tp]
\centering
\footnotesize
\setlength{\tabcolsep}{3pt}
\caption{Percentile requests on minimum TTC for pure car-following histories: 96 evaluation histories (32 per vehicle-count group), five percentiles and three noise draws (1,440 requests, 408 with a target at an atom of the bounded TTC reference, 330 at TTC$_0$ and 78 at the 20~s cap). The upper block uses the generator of the main experiments without retraining. The lower block uses a generator trained only on car-following clips with TTC percentile labels, where Full combines conditioning and guidance. Guidance steers sampling toward the target of the bounded TTC reference. TTC-MAE: mean absolute TTC-target error (s). Cont. and Atom: Fine for requests whose target lies in the continuous part of the reference or at an atom. Bold marks the best value in each column.}
\label{tab:ttc-control}
\begin{tabular}{@{}lrrrrr@{}}
\toprule
Configuration & Fine (\%) & P-MAE & TTC-MAE & Cont. Fine (\%) & Atom Fine (\%)\\
\midrule
\multicolumn{6}{l}{\emph{Generator trained on all clips}}\\
No guidance & 17.22 & 0.30670 & 3.479 & 10.56 & 34.07\\
TTC guidance & 98.61 & 0.00532 & \textbf{0.057} & 98.55 & 98.77\\
\midrule
\multicolumn{6}{l}{\emph{Generator trained on car-following clips}}\\
Full & \textbf{98.82} & \textbf{0.00499} & 0.060 & \textbf{98.64} & \textbf{99.26}\\
Condition only & 21.60 & 0.27589 & 3.184 & 11.24 & 47.79\\
Guidance only & 98.75 & 0.00541 & 0.069 & 98.55 & \textbf{99.26}\\
No P & 20.56 & 0.29144 & 3.393 & 9.88 & 47.55\\
\bottomrule
\end{tabular}
\end{table}
The interface accepts any scalar surrogate with a criticality direction. We test this generality on car following without lane changes, where TTC is the customary surrogate \citep{westhofen2023criticality}. The surrogate is the minimum TTC between the ego and its leader over the future window, for clips in which both keep their lane and no other vehicle enters between them. TTC is capped at 20~s, beyond which an approach has little bearing on safety, and rescaled to the support of the reference. A TTC reference with the architecture of the range-query reference is fitted on the car-following clips of the training recordings (Appendix~\ref{app:ttc-details}). Because the future window starts at the last observed frame, the minimum TTC cannot exceed TTC$_0$, the TTC at the start of the future, which the history fixes. We bound the reference accordingly by moving the mass it places above TTC$_0$ to an atom at TTC$_0$. Every realized future respects this bound, so bounding never raises CRPS. On the car-following clips of the primary set, the bounded reference more than halves the CRPS of the unconditional distribution, from 1.21 to 0.52~s.

Two generators receive the same TTC guidance. The first is the generator of the main experiments, used without retraining through its trained null branch. The second is a TTC version of our method trained only on car-following clips. The guidance moves the largest inverse TTC of the ego and its leader over the future toward the inverse of the reference target, and a target at the atom at TTC$_0$ asks the inverse TTC to stay at or below its initial value. Most car-following histories make an approach unlikely, so the evaluation uses 96 histories whose bounded reference places at most half of its mass at the cap, 32 per vehicle-count group, with the same five requests and three noise draws.

TTC requests are met with a precision similar to that of PET requests (Table~\ref{tab:ttc-control}). With the generator of the main experiments, TTC guidance meets 98.61\% of the requests, against 17.22\% without guidance, and a generator trained only on car-following clips adds little. As for PET, sampling-time guidance carries the precision, and a scalar condition alone gives little control over the realized percentile.

The bound also clarifies what a low request asks for. At $p=0.1$, most targets lie at an atom, so these requests ask for a future in which the approach never becomes more critical than at its start, or in which TTC never falls below the cap. Higher requests ask for closer approaches. The generated futures show how this adversity arises. As the request rises, the leader slows down more, its largest speed drop growing from a median of 0.39 to 1.30~m/s, while the ego brakes less and the minimum TTC moves from the start to the end of the window. High adversity in car following thus builds up late in the window, as the gap to a slowing leader closes. Once the reference respects the bound that the history places on it, the interface transfers to a second surrogate without retraining the generator.

\subsection{Testing a rule-based planner}
\label{sec:planner-test}
\begin{table}[tp]
\centering
\footnotesize
\setlength{\tabcolsep}{3pt}
\caption{An IDM and MOBIL planner tested against generated SV behavior on the 96 evaluation histories. The planner replaces the ego from its observed state at the start of the future and drives for 6.96~s, while the SVs replay their futures without reacting. Our method receives percentile requests, and RADE receives three fixed physical risk levels, given by their target PET. Rear coll.: an SV reaches the planner from behind. Hard brake: planner deceleration beyond 4~m/s$^2$. PET: minimum ego--SV occupancy PET of the planner future. Rear TET: mean time the planner spends with a TTC below 3~s to the SV behind it in its lane \citep{minderhoud2001extended}. Each generated row covers 288 scenarios (96 histories, three noise draws).}
\label{tab:planner-test}
\begin{tabular}{@{}llrrrrr@{}}
\toprule
Scenarios & Request & Rear coll. (\%) & Hard brake (\%) & PET $<1$~s (\%) & Median PET (s) & Rear TET (s)\\
\midrule
Observed futures & -- & 1.04 & 3.12 & 44.79 & 1.128 & 0.048\\
\midrule
Our method & $p=0.1$ & 0.00 & 3.82 & 39.58 & 1.210 & 0.008\\
 & $p=0.3$ & 0.69 & 3.47 & 42.71 & 1.178 & 0.031\\
 & $p=0.5$ & 1.04 & 3.82 & 44.44 & 1.161 & 0.083\\
 & $p=0.7$ & 1.39 & 4.51 & 45.14 & 1.144 & 0.098\\
 & $p=0.9$ & 14.58 & 12.85 & 49.65 & 1.030 & 0.578\\
\midrule
RADE & PET 3.0~s & 1.04 & 4.17 & 44.79 & 1.133 & 0.108\\
 & PET 1.5~s & 4.51 & 6.25 & 46.18 & 1.095 & 0.220\\
 & PET 0.5~s & 2.78 & 8.68 & 50.69 & 0.989 & 0.108\\
\bottomrule
\end{tabular}
\end{table}
A percentile request serves AV testing only if it translates into graded difficulty for the system under test. To check this, we replace the ego of each scenario with the intelligent driver model (IDM) for longitudinal control \citep{treiber2000idm} and MOBIL (minimizing overall braking induced by lane changes) for lane changes \citep{kesting2007mobil}. The planner starts from the observed ego state, its desired time gap is the observed time gap to its leader, bounded to 0.6--1.5~s, and its desired speed is its highest speed in the history. The SVs replay their generated futures without reacting, as the adversarial agents of AdvSim, STRIVE and CAT do when a planner is tested \citep{wang2021advsim,rempe2022strive,zhang2023cat}. Besides collisions, hard braking and PET, we record the time the planner spends with a TTC below 3~s to the SV behind it in its lane, the rear time-exposed TTC (TET) of \citet{minderhoud2001extended}.

The request reaches the planner. Placed in the reference distribution of its own history, the encounter that the planner meets has a median percentile of 0.21 at $p=0.1$ and 0.68 at $p=0.9$, against 0.48 for the replayed observed futures. Because the planner reacts, these percentiles spread more widely than those of the generated futures, but within each history and noise draw they follow the request, with a median Spearman correlation of 0.74. The physical difficulty concentrates at the highest request (Table~\ref{tab:planner-test}). Up to $p=0.7$, hard braking by the planner stays below 5\%, close to the 3.12\% of the observed futures, and at $p=0.9$ it reaches 12.85\% while the rear TET jumps to 0.578~s. RADE's most severe level, a PET target of 0.5~s, reaches a lower median PET than $p=0.9$, 0.989 against 1.030~s, because a fixed physical target pushes every history toward the same severity. Its hard braking rises less, its rear pressure does not grow steadily across its three levels, and its scenes do not occupy a requested position in their own histories (Table~\ref{tab:canonical-control}). All collisions in the scenarios of our method come from SVs that reach the planner from behind, in 14.58\% of the scenarios at $p=0.9$. Because the replayed SVs cannot react, these collisions measure pressure from behind on the planner, which the rear TET captures before any collision.

\begin{figure*}[tp]
\centering
\includegraphics[width=\linewidth]{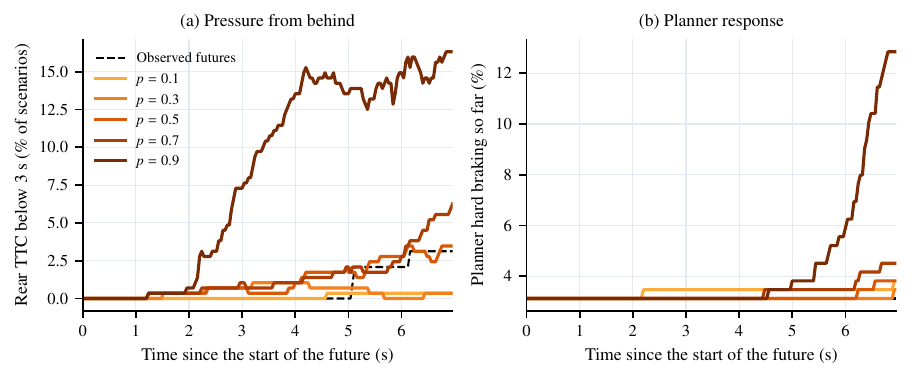}
\caption{How difficulty unfolds in the planner test, for the 288 scenarios of each request and the 96 observed futures. (a) Share of scenarios in which an SV behind the planner in its lane approaches with a TTC below 3~s. (b) Share of scenarios in which the planner has braked harder than 4~m/s$^2$ so far. Hard braking at the start of the window comes from the shared histories.}
\label{fig:planner-dynamics}
\end{figure*}

Fig.~\ref{fig:planner-dynamics} shows when this difficulty arises. At $p=0.9$, SVs begin to approach the planner from behind with a TTC below 3~s after about two seconds, and the share of such scenarios keeps growing until the end of the window, while it stays low for the other requests and for the observed futures. Hard braking by the planner comes late, mostly in the last 1.5~s. High requests thus build up pressure over the window and leave the planner its hardest decisions at the end, as the minimum TTC moved to the end of the window for high TTC requests in Section~\ref{sec:ttc-control}.

\begin{figure*}[tp]
\centering
\includegraphics[width=\linewidth,height=.78\textheight,keepaspectratio]{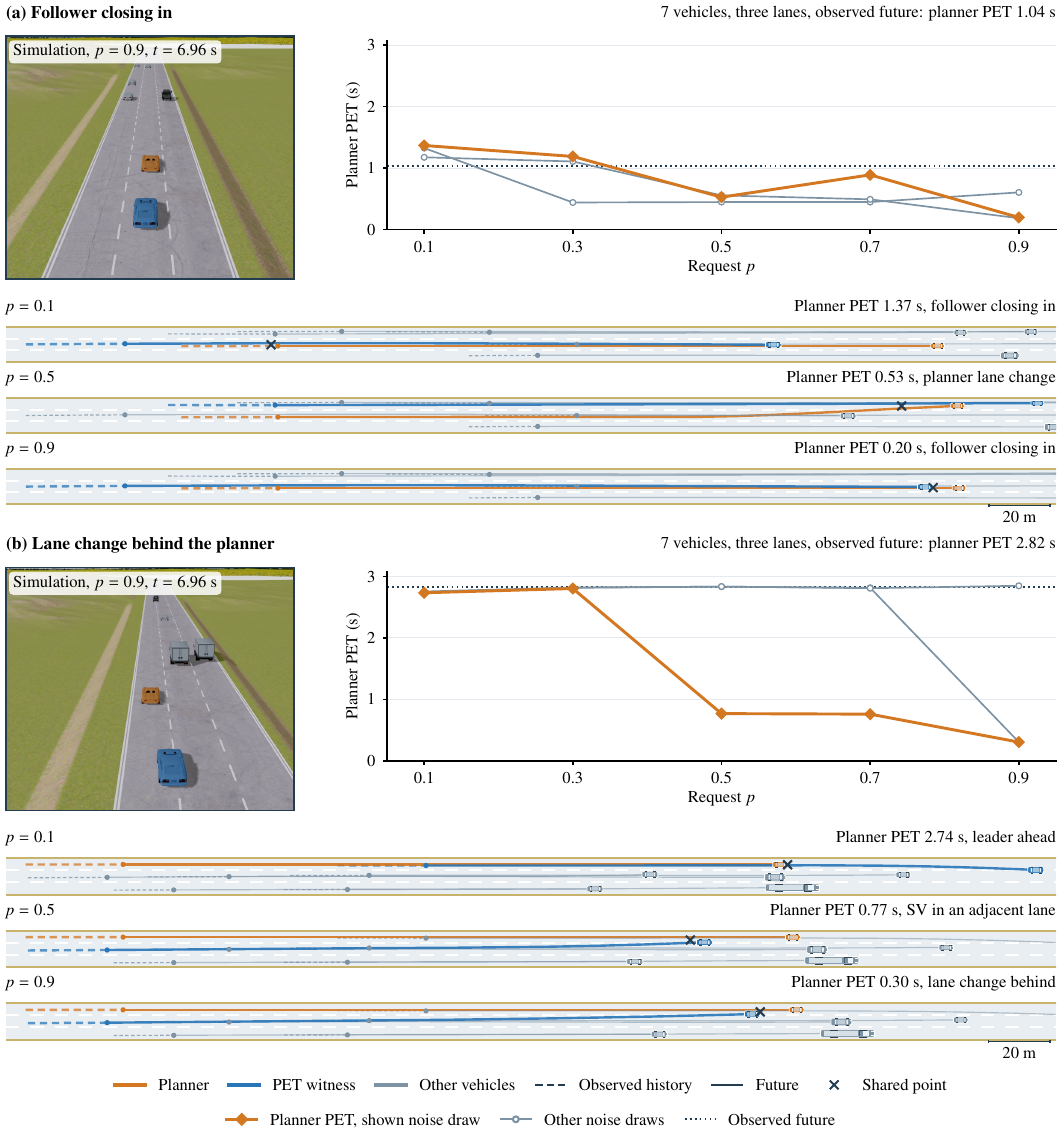}
\caption{Planner test on two histories, drawn as in Fig.~\ref{fig:external-vehicle-scenes}. At $p=0.9$, the encounter that sets the planner's minimum PET is (a) a follower closing in and (b) a lane change behind the planner. The IDM and MOBIL planner replaces the ego, and the SVs replay their generated futures without reacting. Each block shows the history and noise draw whose planner PET falls the most from $p=0.1$ to $p=0.9$ among the collision-free scenarios with this encounter type at $p=0.9$. Top left: simulation view behind the planner at $p=0.9$ when the second of the two vehicles reaches the shared point, with the vehicles colored as in the road views. Top right: planner PET against the request for the three noise draws of the history, with the observed future as a dotted line. Below: road views at $p=0.1$, 0.5 and 0.9. Each row gives the planner's minimum PET and the type of the encounter that sets it, and the cross marks its shared point. Vehicle bodies are drawn at the end of the window and dots at the start of the future. Each block has its own extent, with equal $x$ and $y$ scales.}
\label{fig:planner-cases-ab}
\end{figure*}

\begin{figure*}[tp]
\centering
\includegraphics[width=\linewidth,height=.78\textheight,keepaspectratio]{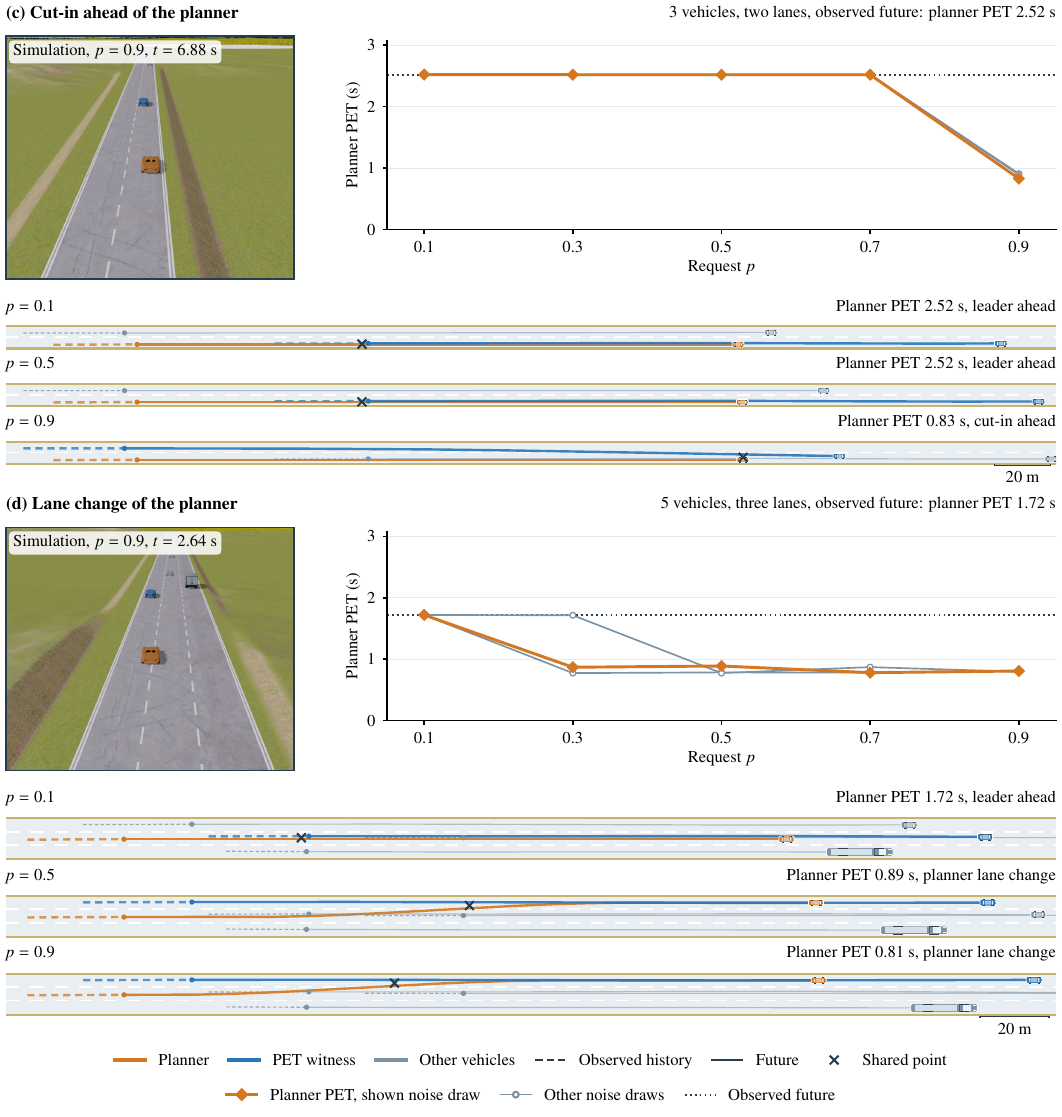}
\caption{Planner test on two further histories, shown as in Fig.~\ref{fig:planner-cases-ab}. At $p=0.9$, the encounter that sets the planner's minimum PET is (c) a cut-in ahead of the planner and (d) a lane change of the planner.}
\label{fig:planner-cases-cd}
\end{figure*}

Figs.~\ref{fig:planner-cases-ab} and~\ref{fig:planner-cases-cd} show four histories in which different types of encounter set the planner's minimum PET at $p=0.9$. An encounter is typed by the lanes of the planner and the SV at the start of the future and at the encounter, and by which of the two passes the shared point first. In Fig.~\ref{fig:planner-cases-ab}(a), the follower in the planner's lane closes in as the request rises. In Fig.~\ref{fig:planner-cases-ab}(b), the encounter moves from a leader far ahead at $p=0.1$ to a vehicle that changes into the planner's lane behind it at $p=0.9$, and the minimum PET falls from 2.74 to 0.30~s. In Fig.~\ref{fig:planner-cases-cd}(c), an SV cuts in ahead of the planner only at $p=0.9$, and in Fig.~\ref{fig:planner-cases-cd}(d) the planner itself changes lanes behind an SV at the higher requests. Within each block the history and the planner are the same, so the request alone sets how hard the encounter becomes. The simulation views, rendered with MetaDrive \citep{li2023metadrive}, show the encounter at $p=0.9$ from behind the planner, and videos of the planner test are available on the project website (\projecturl). The PET panels show the same trend for nearly all noise draws of these histories.

Across all scenarios, the type of encounter that sets the planner's minimum PET changes little with the request, and a leader ahead remains the most common type. With a leader ahead, the planner keeps its own time gap. Where a leader ahead sets the minimum PET at both $p=0.1$ and $p=0.9$, the median planner PET is 1.53~s at both requests, whereas for a follower closing in it falls from 1.79 to 0.86~s. The request thus reaches the planner mainly through the vehicles behind and beside it.

For a test designer, the request therefore places a scenario on the scale of its own history, and its strongest physical effect appears at the highest request. At $p=0.9$, the planner meets hard braking 4.1 times as often as in the observed futures of the same histories, and $p=0.1$ moves the encounter to the calm side of each history.

\section{Implications for scenario-based AV testing}
\label{sec:implications}
For AV testing, the percentile states how corner a generated scenario is relative to the natural futures of its own history. Its physical target adapts to that distribution, giving the same numerical request a common relative interpretation across traffic contexts. A corner-case library can therefore be indexed by history and adversity percentile, with each entry also recording its realized surrogate value. Because each physical target is known before generation, a test that also requires a physical severity, such as a PET below one second, can select the histories whose targets reach it. Holding $H$ and $z$ fixed fine-tunes cornerness through paired sweeps over $p$.

This request definition connects distribution estimation, generation and evaluation through one reference. Its inverse determines the physical target used by generation, and its compatible percentile interval determines how the returned outcome is scored. CRPS and calibration diagnostics assess the estimated natural distribution. Percentile and physical-target errors assess request realization under that estimate. Updating the reference consequently requires recomputing percentile labels, CDF conditions and physical targets together, and adapting the generator to them.

The interface takes a scalar surrogate and its criticality direction as inputs. A further instantiation would estimate that surrogate's history-conditioned natural distribution and implement a mechanism for realizing its inverse-CDF targets. The present experiments establish this chain for minimum ego--SV PET: the reference supplies the target, conditioning supplies a request-dependent proposal, and sampling-time guidance provides most of the final precision. The occupancy-geometry updates implement PET-target realization specifically.

The evidence has limits. The study addresses highway scenarios and uses highD throughout, so other highway datasets and further surrogates remain to be tested. A percentile is as accurate as the reference that defines it, which is why reference accuracy and request realization are reported separately. Each model was trained once, so the reported intervals describe variation across histories and recordings for one trained model. The planner test replays SVs that do not react, which holds the scenario fixed while the planner responds.

The request definition leaves room for several extensions. The most direct one covers further scenarios, such as merging at highway entries and exits and crossing conflicts at urban intersections, where PET is a common surrogate \citep{fhwa2008surrogate}. Each needs a reference fitted on its own natural data and a realization mechanism for its encounter geometry, while the request and its evaluation stay the same. Requests could also combine several surrogates, or a percentile with a semantic request such as a lane change \citep{tan2023lctgen}, given a reference conditioned on both. Reactive SVs would make the planner test closed-loop, with SV futures generated online so that the encounter stays at the requested percentile as the planner responds. A test campaign could choose its next requests from earlier outcomes, as adaptive testing chooses scenarios \citep{yang2025adaptive}. Since the percentile of a natural future is uniformly distributed under a calibrated reference, results over a grid of requests could also be combined into an estimate of how often a planner fails in natural traffic, provided the generated futures follow natural futures at each risk level.

\section{Conclusions}
Corner-case scenarios for AV testing need a calibrated account of how extreme they are. We represent the adversity of a generated scenario as its percentile in the history-conditioned distribution of future risk and make this percentile the generation request. A learned reference risk distribution maps each request to a physical risk target, a percentile-conditioned joint diffusion model with sampling-time risk guidance realizes that target, and a reference-based criterion evaluates the realized percentile consistently at continuous outcomes and probability masses.

The experiments lead to four findings.
\begin{enumerate}
\item With minimum ego--SV PET as the surrogate, our method realizes 1,422 of the 1,440 requests in the primary evaluation set within the 0.05 percentile tolerance, with mean percentile error 0.00673 and PET-target error 0.00991 seconds.
\item The decomposed evaluation assesses reference prediction and request realization separately and identifies guidance as the main source of fine precision, with an additional gain from learned conditioning. A PET-derived risk level, as in RADE, shifts the overall severity of the scenes yet controls the contextual percentile no better than learned conditioning on the percentile.
\item The interface carries over to TTC requests in car following without retraining the generator, and higher requests pose harder scenarios to an IDM and MOBIL planner.
\item Across the three evaluation sets, the futures generated by our method contain no footprint overlap between background vehicles.
\end{enumerate}

The contribution lies in this request definition. The reference estimator and the generator used here are one implementation of it, and the ablation shows that requests are realized once sampling is steered toward the physical target that the reference derives from them. The resulting interface connects context-relative risk specification, physical realization and evaluation through a common risk scale. Section~\ref{sec:implications} discusses the limits of the present evidence and the extensions that the request definition allows.

\label{sec:main-end}

\section*{CRediT authorship contribution statement}
\textbf{Jiaxi~Liu:} Writing -- review \& editing, Writing -- original draft, Visualization, Validation, Software, Methodology, Investigation, Formal analysis, Data curation, Conceptualization. \textbf{Hang~Zhou:} Writing -- review \& editing, Methodology. \textbf{Hangyu~Li:} Writing -- review \& editing. \textbf{Yifan~Wang:} Writing -- review \& editing. \textbf{Keke~Long:} Writing -- review \& editing. \textbf{Chengyuan~Ma:} Writing -- review \& editing. \textbf{Bin~Ran:} Supervision. \textbf{Xiaopeng~Li:} Writing -- review \& editing, Supervision, Conceptualization.

\section*{Declaration of competing interest}
The authors declare that they have no known competing financial interests or personal relationships that could have appeared to influence the work reported in this paper.

\section*{Acknowledgements}
This material is based upon work supported by the U.S. National Science Foundation (NSF) under the NSF-DST Cyber-Physical Systems program, Award No. 2343167.

\section*{Data availability}
This study uses the raw highD dataset \citep{krajewski2018highd}, which its provider makes freely available for non-commercial research upon request under the dataset's access policy (\url{https://levelxdata.com/highd-dataset}). The code and the pretrained models are available at \url{https://github.com/hhjj233/P-control-SG}.

\appendix
\section{Implementation details}
\label{app:canonical-details}

\subsection{Clips, vehicles and visibility}
\label{app:canonical-data}
A clip is defined by a recording, an ego vehicle and an anchor frame. Anchors lie on a 25-frame grid, and the 13 historical observations are sampled every two native frames, spanning 0.96 seconds at 25 Hz. At the anchor, the scene contains the ego and every observed car or truck that travels on the same side of the road in the same direction, lies within 120 longitudinal meters and has a complete history. These vehicles are defined from the history alone, at least three including the ego, and their number varies with traffic. In each recording, candidate histories are drawn up to a fixed number in an order that does not depend on their futures.

The future contains 175 native states, including the initial state, over 6.96 seconds. Let $E=1$ mean that every vehicle of the scene is observed at all those times. The chosen reference population in Eq.~\eqref{eq:population} is consequently
\[
F^\star_H(y)=\Prob\{Y(X,H)\le y\mid H,E=1\}.
\]
The event $E$ defines the reference population. It is not a model input, and it does not guarantee that the vehicles remain observable at deployment. Clips with $E=0$ are excluded without replacement. They are not completed by deleting missing vehicles, extrapolating motion or inserting simulated futures. Their observed minimum PET only bounds the minimum over all vehicles from above, so it cannot serve as an ordinary CRPS target.

The 13 training recordings provide 53,248 candidate histories, of which 9,913 have $E=1$. These training clips contain 3--37 vehicles. Separate recordings supply the early-stopping set (538 clips), the calibration set (504 clips) and a development set on which model variants are compared. Model weights and calibration maps are selected on the early-stopping and calibration sets. The three evaluation sets of Section~\ref{sec:experiments} are excluded from fitting, model selection and calibration. Normalization statistics come from the training clips only. Model inputs contain history, dimensions, road bounds and valid-slot masks, ego indicators and actor masks. Future visibility, recording identity, vehicle identifiers and measured future PET are not inputs.

The PET surrogate in Eq.~\eqref{eq:scene-pet} uses closed, axis-aligned footprints with measured lengths and widths, with positions interpolated linearly between adjacent native frames, and its minimum over all pairs of the ego and an SV is computed exactly. Both occupation times are restricted to the observed or generated 6.96-second window, without future extrapolation. Closed-edge contact counts as shared occupancy, positive PET is not rounded to zero, and outcomes close to the four-second cap are not moved onto it.

\subsection{Reference architecture, training and calibration}
\label{app:canonical-reference}
Absolute and ego-relative temporal encoders summarize each vehicle's history, relation tokens pair the ego with each SV, and PET-range queries read these tokens for the components of the mixed CDF. Each of the two actor-shared temporal encoders uses a learned summary token, learned temporal positions, two Transformer layers, four attention heads, width 64, feed-forward width 128, GELU and zero dropout. The absolute and ego-relative streams have separate parameters. All temporal inputs precede the generation time, so attention within the historical window is unrestricted. Dimensions are projected and fused with the absolute summary before normalization. Road-boundary pooling and $\log(1+N)$ form the scene context. A separate ego-context token supplements the ego--SV relation set.

The reference has 237,633 parameters. Its relation, road/count and output mappings include MLPs. The Transformer layers form the two history encoders. There are 64 continuous PET bins and two endpoint atoms. Each query is specified by its component's fixed interval bounds and type. The static-query variant keeps the same trainable tensor shapes and range-aware output scorer, but reads the relations with one range-independent query. The two readouts therefore have matched parameter counts. The range-query readout evaluates 66 queries and uses more attention computation.

Training minimizes ordinary CRPS with AdamW (learning rate $10^{-3}$, weight decay $10^{-4}$, batch 128, gradient clipping 5) for at most 80 epochs with early-stopping patience 12, and the weights with the lowest overall CRPS on the early-stopping set are kept.

The calibration warp is piecewise linear, with probability knots
\[
0,\ 0.05,\ 0.1,\ 0.25,\ 0.5,\ 0.75,\ 0.9,\ 0.95,\ 1.
\]
Its endpoints are fixed at zero and one. The count-conditioned family mixes two monotone maps with weight $\operatorname{clip}((N-5)/8,0,1)$. Calibration minimizes normalized CRPS on the calibration set, with ridge shrinkage toward the identity map. Leave-one-recording-out validation within the calibration set selects among the identity map, a global warp and the count-conditioned warp at ridge strengths from $10^{-4}$ to 1, with safeguards on overall, group and tail-weighted CRPS \citep{allen2023weighted}. The selected map is count-conditioned with ridge strength 1.

Composing the uncalibrated CDF with its warp changes the endpoint masses and can add physical breakpoints, and the composed distribution keeps both. Left and right CDF values, inverse targets, ranks and CRPS all use it. The generator receives a 65-value shape descriptor sampled on a physical PET grid, with the left limit used at the cap.

\subsection{Leave-recording-out percentile labels}
\label{app:canonical-oof}
Observed PET supervises the reference. Fitted references then assign the percentile labels used to train the generator. To keep each labeled clip outside the training data of the reference that labels it, the 13 training recordings are divided into five groups. For each group, a reference with its own normalizer is trained on the other four groups for 47 epochs of ordinary CRPS training, the length at which the full reference was early-stopped, and a count-conditioned map with ridge strength 1 is fitted on the calibration set. We call these five models the leave-recording-out references.

For a clip $i$ from recording $r_i$, let $\G^{(-r_i)}_{H_i}$ denote the leave-recording-out reference whose training data exclude $r_i$. The clip's out-of-fold label is
\begin{equation}
p_i^{\rm OOF}
=1-\tfrac12\left[
\G^{(-r_i)}_{H_i}(y_i^-)+\G^{(-r_i)}_{H_i}(y_i)\right].
\label{eq:oof}
\end{equation}
At an atom, this label is the center of the compatible interval, whereas evaluation scores a request against the whole interval (Appendix~\ref{app:canonical-evaluation}).

Each training clip's percentile label, CDF descriptor and inverse target come from the same leave-recording-out reference. Early-stopping clips and all generation requests use the reference fitted on all training recordings, with its own calibration map. Because each leave-recording-out reference is fitted on fewer recordings, its estimate for a given history can differ from that of the full reference.

\subsection{Generator and sampling details}
\label{app:canonical-training}
The future decoder uses eight acceleration-cosine modes per coordinate. In Eq.~\eqref{eq:decoder}, $B_0(t)=t^2/2$ and $B_k(t)=(1-\cos(\omega_kt))/\omega_k^2$ for $k>0$, with $\omega_k=k\pi/6.96$. Least squares fits observed natural positions to this representation, introducing a coefficient-reconstruction approximation. Initial positions and velocities are fixed by the decoder.

The denoiser has width 128, four heads, three blocks, feed-forward width 256 and 980,368 parameters including conditioning adapters. It uses a 100-step cosine noise schedule, masked $v$-prediction MSE and 50-step deterministic DDIM. Actor tokens contain history, dimensions, ego role, road context, noisy coefficients and diffusion time. The risk branch receives $p$, $q_H(p)/\Tcap$ and the CDF descriptor. Per-block residual adapters allow actor-dependent responses to the shared request.

The trained null-risk branch masks the entire learned risk-conditioning pathway: the percentile modulation, the target- and CDF-dependent scale and shift, and the dynamic adapters. History, road, dimension, ego and count information remain present.

The generator is trained in stages on the natural training clips, each stage continuing from the weights of the previous one. The code repository lists every stage with its settings (see Data availability). A natural adaptation stage to the final reference precedes the final risk-aware stage. The final stage combines the natural diffusion loss with a direction term and a condition-response term, both built from natural coefficient tangents and the reference CDF density, and with penalties on the terminal samples of auxiliary 50-step rollouts at $p=0.1,0.5,0.9$. These penalties cover the percentile and physical PET errors with respect to the targets, a fine-band term, an encounter term, road violations and reverse speed. Gradients pass through the full rollout. The released generator uses the weights after the third epoch of this stage without averaging.

Risk guidance is enabled in the final 15 DDIM steps. Its preconditioner weights the position, velocity and acceleration basis metrics by 1, 0.25 and 0.0625, with eigenvalues floored at $10^{-4}$ of the largest. Each proposed direction is checked with a probe of at most $10^{-4}$ meters and a complete feasible PET query, and updates stop once the center-rank error is at most 0.01.

Road projection minimizes metric displacement under outer-road footprint inequalities, with an inward margin of up to 0.002 meters and each actor's initial allowance kept unchanged. Background quadratic programs alter only longitudinal coefficients, fixing the ego and all lateral coefficients, in up to four passes that add newly active pairs. Where the observed initial state already has insufficient clearance, it is kept as observed. Risk proposals pass through these constraints, and the geometry projections are not subject to the 0.5-meter risk-increment bound.

Rare strong accelerations remain in the futures of our method. Accelerations above 8~m/s$^2$ in magnitude occur in 0.09\% of the vehicle-time samples, against 0.01\% in the observed futures (Table~\ref{tab:kinematic-realism}). They concentrate on the ego, which takes part in every encounter that guidance adjusts, with 0.45\% of its samples, and on low requests, with 0.23\% of the samples at $p=0.1$ against 0.01\% at $p=0.7$.

\subsection{Time-to-collision requests}
\label{app:ttc-details}
The TTC reference is fitted on 6,356 car-following clips from 11 of the 13 training recordings, with 752 clips from the other two for early stopping. The TTC generator uses the percentile-conditioned architecture of the conditioning pathway and cross-fitted percentile labels from the bounded reference. Guidance acts in the last 30 sampling steps. Its direction follows a log-sum-exp smooth maximum of the inverse TTC over time with $\beta=200$~s, so one step lowers every instant near the maximum, and its step size follows the exact maximum. For a target at the atom at TTC$_0$, the guidance aims 0.002~s$^{-1}$ below the initial inverse TTC. The cap and the guidance settings were chosen on the two early-stopping recordings. Of the 7,712 car-following clips of the primary set, 1,240 have a bounded reference with at most half of its mass at the cap, and the evaluation histories are drawn from them.

\subsection{Interval-compatible evaluation}
\label{app:canonical-evaluation}
For $I_H(y)=[L_H(y),U_H(y)]$, interval error is computed without randomizing a percentile within the atom:
\[
e_{\rm int}=\max\{L_H(y)-p,\ p-U_H(y),\ 0\}.
\]
At a continuous outcome this equals the ordinary percentile error, and at an atom it recognizes all requests compatible with the same physical outcome. Equation~\eqref{eq:quantile-compatibility} follows from the definition of the generalized inverse and the right-continuity of the CDF, so $Y=q_H(p)$ implies $e_{\rm int}=0$.

The deterministic center rank
\[
r_H(y)=1-\tfrac12\{\G_H(y^-)+\G_H(y)\}
\]
assigns a single percentile to each outcome. It is the training label of Eq.~\eqref{eq:oof} and enters the percentile penalty of the final training stage and the stopping rule of the sampler (Appendix~\ref{app:canonical-training}). Evaluation uses the whole compatible interval, because minimum PET has point masses at zero and at the four-second cap: at an atom, every percentile in the interval is realized by the same physical outcome. Target types are defined by $q_H(p)$ before generation.

For completeness, the reference and realization errors can be separated for the interval criterion. For a CDF $F$, define $I_F(y)=[1-F(y),1-F(y^-)]$, and let
\[
\epsilon_H=\sup_y|\G_H(y)-F_H^\star(y)|.
\]
The same bound holds for left limits. The two endpoints of $I_{\G_H}(y)$ and $I_{F_H^\star}(y)$ therefore differ by at most $\epsilon_H$, so the Hausdorff distance between these intervals is at most $\epsilon_H$. Distance to an interval is Lipschitz with respect to that distance, yielding
\begin{equation}
\operatorname{dist}(p,I_{F_H^\star}(y))
\le \operatorname{dist}(p,I_{\G_H}(y))+\epsilon_H.
\label{eq:reference-control-error}
\end{equation}
This relation identifies the two quantities assessed by the reference and generation evaluations. Estimating $\epsilon_H$ for an individual history requires history-specific information beyond the aggregate CRPS and pooled PIT diagnostics reported here.

\section{Reference encoders and calibration diagnostics}
\label{app:context-diagnostics}
\label{app:method-details}
The reference of Section~\ref{sec:problem} can be estimated with different encoders. Table~\ref{tab:final-reference-encoders} compares six of them, trained and calibrated with the same protocol: a parameter-matched wide MLP, two gated recurrent unit (GRU) streams, and Transformer encoders that read absolute histories only, ego-relative histories only, or both. All use the range-query readout except the static-query reference, which replaces only this readout by its range-independent counterpart.

\begin{table}[t]
\centering
\small
\caption{Reference encoders on the primary evaluation set (11,649 clips from 16 recordings). CRPS (seconds) of the calibrated models, with PIT$_{\rm tail}$ and PIT$_{\max}$ defined in the text. The range-query reference is the one used in all experiments (ours), and the static-query reference reads the same histories with one static query. All other encoders use range queries. Lower is better in every column. Bold marks the lowest value in each column.}
\label{tab:final-reference-encoders}
\begin{tabular}{@{}lrrr@{}}
\toprule
Model & CRPS & PIT$_{\rm tail}$ & PIT$_{\max}$\\
\midrule
Wide MLP & 0.10846 & 0.03490 & 0.08911\\
GRU & 0.11375 & 0.03767 & 0.11007\\
Absolute histories only & \textbf{0.10349} & 0.01996 & \textbf{0.03342}\\
Relative histories only & 0.11631 & 0.03294 & 0.05560\\
Static-query reference & 0.11160 & 0.04318 & 0.05024\\
Range-query reference (ours) & 0.10550 & \textbf{0.01051} & 0.07915\\
\bottomrule
\end{tabular}
\end{table}
All neural encoders reach a calibrated CRPS between 0.103 and 0.116 seconds, three to six times below the classical estimators of Table~\ref{tab:canonical-reference}, so the choice among them changes the reference little. The encoder that reads absolute histories only is the most accurate by a small margin. The range-query reference used in our experiments has the lowest CRPS on the development set, 0.0991 seconds after calibration against at least 0.1125 for the other encoders, and the smallest deviation in the critical tail, while its largest deviation lies on the benign side. On the additional and single-recording sets, it again ranks second in CRPS, and its tail deviation stays at or below 0.018, against 0.054 and 0.064 for the static-query reference. Any of these encoders could supply the reference, since a request depends only on the calibrated distribution that the reference produces.

For mixed distributions, randomized PIT is $U=\G_H(y^-)+V\{\G_H(y)-\G_H(y^-)\}$ with independent $V\sim\mathrm{Unif}(0,1)$. Expected-PIT diagnostics integrate this randomization analytically at atoms. PIT$_{\max}$ is the largest deviation of the expected-PIT CDF from the diagonal over the grid $0.05,0.10,\ldots,0.95$, and PIT$_{\rm tail}$ restricts it to the levels 0.05--0.30, where the targets of requests $p\ge0.7$ lie. These grid errors differ from a Kolmogorov--Smirnov statistic of sampled PIT values.

\section{External components without percentile requests}
\label{app:external-components}
The three external models are trained on the same training clips, with model selection on the early-stopping set. At generation they receive only the observed history and static scene information. The reference is used afterwards to score their outputs. Every vehicle of the history is generated. Native time grids are mapped deterministically to the common 175-state, 6.96-second evaluation grid, keeping the observed initial state exactly.

TrafficGen uses its motion actuator and loss and starts from the observed initial scene \citep{feng2022trafficgen}. Its placement stage is not used. Every vehicle is predicted in its current local frame with the full initial scene as context, without the observed ego future and without truncating the scene to 32 agents. It predicts 58 steps of 0.12 seconds with its native top-1 mode decoder. With fixed initial states this output is deterministic, so TrafficGen contributes one output per history (96 in total).

CTG++ uses its unguided scene-diffusion prior with its unicycle dynamics, 58 steps of 0.12 seconds and 100 denoising steps, without language or editing guidance \citep{ctgpp2023}. STRIVE uses its graph-CVAE traffic prior and bicycle decoder, with 29 steps of 0.24 seconds and position- and orientation-dependent rasters built from the observed road boundaries, without its adversarial planner optimization \citep{rempe2022strive}. CTG++ and STRIVE each draw 15 samples per history.

STRIVE produces many strict road flags, most of them in scenes that start compliant, and many remain with a 0.5-meter tolerance. Its native road loss supervises only the ego, whereas the road check covers all vehicles.

The comparison thus concerns these motion-generation components as adapted here, each with its own training and inference budget.

\begin{figure}[tp]
\centering
\includegraphics[width=\linewidth,height=.80\textheight,keepaspectratio]{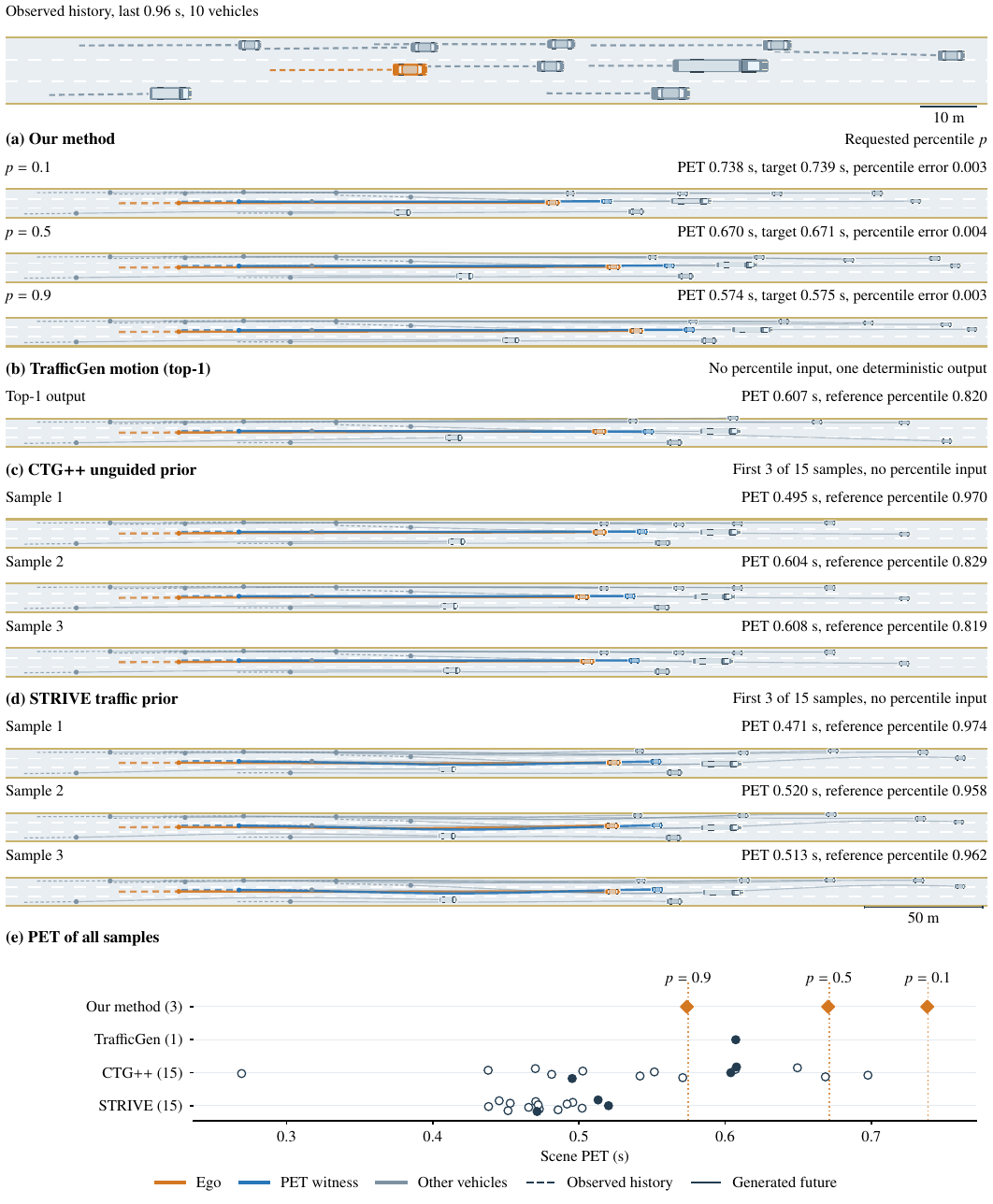}
\caption{External priors on the car-following history of Fig.~\ref{fig:qualitative-scenes}(c), without post-selection. The top view shows the observed history at a larger scale. (a) Our method at the requested $p=0.1,0.5,0.9$, as in Fig.~\ref{fig:external-vehicle-scenes}. (b) TrafficGen motion, top-1 output without a percentile input. (c) CTG++ unguided prior and (d) STRIVE traffic prior, each showing the first three of its 15 samples. Each row reports the scene PET with the target and percentile error (our method) or the reference percentile of the sample (priors). (e) PET of all samples (Our method 3, TrafficGen 1, CTG++ 15, STRIVE 15), with filled markers for the displayed samples and dotted lines at the targets.}
\label{fig:external-vehicle-noise-zero}
\end{figure}

\paragraph{Same-history visual comparisons}
Fig.~\ref{fig:external-vehicle-scenes} and Fig.~\ref{fig:external-vehicle-noise-zero} use the ten-vehicle car-following history of Fig.~\ref{fig:qualitative-scenes}(c) and show the same three futures generated by our method at $p=0.1,0.5,0.9$. Fig.~\ref{fig:external-vehicle-noise-zero} shows the first three of the 15 CTG++ and STRIVE samples, whereas Fig.~\ref{fig:external-vehicle-scenes} shows the samples chosen by the post-selection rule below. TrafficGen has its single top-1 output in both. Both figures draw on the same 15 samples per method.

\paragraph{Post-selection for the main external visualization}
\label{app:external-visual-postselection}
For each $p\in\{0.1,0.5,0.9\}$, we select independently among the 15 samples of an external method, which were generated without a percentile input:
\[
k^\star(p)=\operatorname*{arg\,min}_{k\in\{1,\ldots,15\}}
\big|Y(X_k,H)-q_H(p)\big|.
\]
Here $Y$ is the physical scene PET. The same sample may be selected for different targets. On this history, CTG++ selects three distinct samples, whereas one STRIVE sample is the closest to all three targets and is shown in each of their rows. The selection uses realized outcomes and serves display only: our method produces each output from a single sampling path ($K=1$), and the aggregate external results use all samples.

The PET panels of both figures show all 15 CTG++ and all 15 STRIVE samples for this history, with filled markers for the displayed samples and open markers for the others. Some prior samples lie close to a target. Showing the complete sets next to the displayed trajectories separates what prior sampling produces from what target-aware selection adds. Scene panels plot the complete 0--6.96-second trajectories, with vehicle bodies at the final state and dots at the initial positions. Panels of one history share an equal-aspect extent, and the history view is drawn at a larger scale. The PET witness is determined separately for each displayed future. In the PET panels, markers are offset only vertically to separate overlapping points.

\paragraph{Selection of the lane-change and car-following examples}
\label{app:generated-lane-change-selection}
The lane-change examples are chosen from the 3,930 generated futures for their maneuvers. A vehicle completes a lane change when its full footprint starts inside one lane and ends inside an adjacent lane, its lateral displacement is at least 1.8 meters, and its whole body stays in the destination lane throughout the final 14 states (0.52 seconds). A crossing of an internal lane boundary by the vehicle's center is counted separately, and leaving through an outer road boundary does not count. A witness lane change must be made by the PET witness of that output. Table~\ref{tab:ego-witness-lane-scan} reports how often these maneuvers occur. All completed ego lane changes occur in the additional and single-recording sets. The requests specify only a risk percentile, so the maneuver is an outcome.
\begin{table}[tp]
\centering\small
\caption{Lane changes of the ego and of the PET witness in futures generated by our method. Center: the vehicle's center crosses an internal lane boundary. Complete: a completed lane change as defined in the text. The PET witness is determined separately for each output.}
\label{tab:ego-witness-lane-scan}
\begin{tabular}{lrrrrr}
\toprule
Evaluation set & Outputs & \shortstack{Ego\\center} & \shortstack{Ego\\complete} & \shortstack{Witness\\center} & \shortstack{Witness\\complete}\\
\midrule
Primary & 1440 & 7 & 0 & 74 & 10\\
Additional & 1440 & 31 & 17 & 70 & 17\\
Single recording & 1050 & 25 & 8 & 63 & 9\\
\bottomrule
\end{tabular}
\end{table}

Each example shows one history and one noise draw at $p=0.1,0.5,0.9$. Lane-change candidates are ranked first by whether the ego or the PET witness completes a lane change at all three requests, and then at any request. Within a rank, primary-set histories come first, in an order that does not depend on the outputs. Two history--noise combinations complete a lane change at all three requests. The first, from the additional set, is the ego lane-change example, in which the ego changes lane at every request. For the PET-witness lane-change example, the search is restricted to primary-set histories with at least six vehicles and a completed PET-witness lane change at one or more requests. Neither lane-change selection uses request error, contacts or the outputs of other methods. In the witness example, only the $p=0.9$ witness changes lane. The $p=0.1$ and 0.5 rows complete the request sweep.

The car-following example comes from the primary set, on which the external models were also run. An output is car following when the ego and its PET witness keep the same lane throughout the future window, with no center crossing and the same start and end lane, and the witness stays ahead of the ego in the driving direction. Among the history--noise combinations with at least six vehicles whose outputs at $p=0.1,0.5,0.9$ are car following with the same witness, and in which both the ego and its witness are cars shorter than 7 meters, the displayed example has the smallest maximum interval error.

\clearpage

\bibliographystyle{cas-model2-names}
\bibliography{reference}
\end{document}